\documentclass{article} 
\usepackage{iclr2025_conference,times}
\usepackage{CJKutf8}
\usepackage{hyperref}
\usepackage{verbatim}
\usepackage[utf8]{inputenc} 
\usepackage[T1]{fontenc}    
\usepackage{url}            
\usepackage{booktabs}       
\usepackage{amsfonts}       
\usepackage{nicefrac}       
\usepackage{microtype}      
\usepackage{xcolor}         

\usepackage{booktabs, multirow} 
\usepackage{soul}
\usepackage{graphicx}
\usepackage{wrapfig}
\usepackage{subcaption}
\usepackage{adjustbox}
\usepackage[most]{tcolorbox}
\usepackage{colortbl}
\usepackage{tikz}
\usepackage{fontawesome5}
\usepackage{makecell}
\usepackage{tablefootnote}

\usepackage{amsmath,amsfonts,bm}

\def\eqref#1{equation~\ref{#1}}

\def\1{\bm{1}}

\DeclareMathAlphabet{\mathsfit}{\encodingdefault}{\sfdefault}{m}{sl}
\SetMathAlphabet{\mathsfit}{bold}{\encodingdefault}{\sfdefault}{bx}{n}

\newcommand{\dataname}[0]{\textsc{Magpie}}

\newtcolorbox{prompt}[2][]{
    colback=white,
    colframe=gray!45,
    fonttitle=\bfseries,
    coltitle=black,
    sharp corners,
    title=#2,
    #1
}

\tcbset{
    promptstyle/.style={
        enhanced,
        colback=white,
        colframe=black,
        colbacktitle=gray!20,
        coltitle=black,
        rounded corners,
        sharp corners=north,
        boxrule=0.5pt,
        drop shadow=black!50!white,
        attach boxed title to top left={
            xshift=-2mm,
            yshift=-2mm
        },
        boxed title style={
            rounded corners,
            size=small,
            colback=gray!20
        }
    },
    replystyleg/.style={
        enhanced,
        colback=green!15,
        colframe=black,
        colbacktitle=green!30,
        coltitle=black,
        boxrule=0.5pt,
        drop shadow=black!50!white,
        rounded corners,
        sharp corners=north,
        attach boxed title to top right={
            xshift=-2mm,
            yshift=-2mm
        },
        boxed title style={
            rounded corners,
            size=small,
            colback=green!40
        }
    },
    replystyler/.style={
        enhanced,
        colback=red!15,
        colframe=black,
        colbacktitle=red!40,
        coltitle=black,
        boxrule=0.5pt,
        drop shadow=black!50!white,
        rounded corners,
        sharp corners=north,
        attach boxed title to top right={
            xshift=-2mm,
            yshift=-2mm
        },
        boxed title style={
            rounded corners,
            size=small,
            colback=red!40
        }
    }
}

\newtcolorbox{promptbox}[1][]{
    promptstyle,
    title=Prompt,
    #1
}

\title{
\fontsize{16.9}{20}\selectfont
\dataname{}: Alignment Data Synthesis from Scratch by Prompting Aligned LLMs with Nothing}

\newcommand{\github}{\raisebox{-1.5pt}{\includegraphics[height=1em]{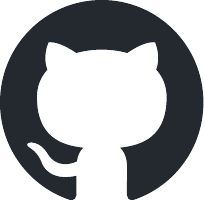}}}
\newcommand{\huggingface}{\raisebox{-1.5pt}{\includegraphics[height=1em]{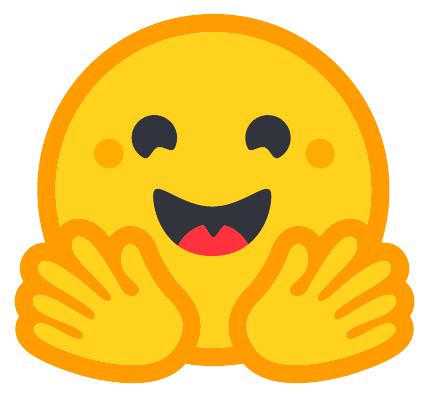}}}

\author{Zhangchen Xu$^\spadesuit$ \qquad Fengqing Jiang $^\spadesuit$ \qquad  Luyao Niu$^\spadesuit$
\qquad  Yuntian Deng $^\diamondsuit$ ~\vspace{0.4em} \\
\textbf{Radha Poovendran$^\spadesuit$ \qquad  Yejin Choi$^\spadesuit$$^\diamondsuit$ \qquad  Bill Yuchen Lin$^\diamondsuit$}
~\vspace{0.5em}\\
$^\spadesuit$University of Washington \qquad
$^\diamondsuit$Allen Institute for AI \vspace{0.5em}
\\
{\small{\github{} {\texttt{\url{https://magpie-align.github.io/}}}}} \\
{\small{\huggingface{} {\texttt{\url{https://hf.co/magpie-align}}} 
}}}

\iclrfinalcopy 
\begin{document}

\maketitle

\vspace{-2em}
\begin{abstract}

High-quality instruction data is critical for aligning large language models (LLMs). Although some models, such as Llama-3-Instruct, have open weights, their alignment data remain private, which hinders the democratization of AI. 
High human labor costs and a limited, predefined scope for prompting prevent existing open-source data creation methods from scaling effectively, potentially limiting the diversity and quality of public alignment datasets.
Is it possible to synthesize high-quality instruction data at scale by extracting it directly from an aligned LLM?
We present a \textit{self-synthesis} method for generating large-scale alignment data named \dataname{}. 
Our key observation is that aligned LLMs like Llama-3-Instruct can generate a user query when we input only the pre-query templates up to the position reserved for user messages, thanks to their auto-regressive nature. 
We use this method to prompt Llama-3-Instruct and generate 4 million instructions along with their corresponding responses. We further introduce extensions of \dataname{} for filtering, generating multi-turn, preference optimization, domain-specific and multilingual datasets.
We perform a comprehensive analysis of the \dataname{}-generated data. To compare \dataname{}-generated data with other public instruction datasets (e.g., ShareGPT, WildChat, Evol-Instruct, UltraChat, OpenHermes, Tulu-V2-Mix, GenQA), we fine-tune Llama-3-8B-Base with each dataset and evaluate the performance of the fine-tuned models. Our results indicate that using \dataname{} for supervised fine-tuning (SFT) solely can surpass the performance of previous public datasets utilized for both SFT and preference optimization, such as direct preference optimization with UltraFeedback. We also show that in some tasks, models supervised fine-tuned with \dataname{} perform comparably to the official Llama-3-8B-Instruct, despite the latter being enhanced with 10 million data points through SFT and subsequent preference optimization. This advantage is evident on alignment benchmarks such as AlpacaEval, ArenaHard, and WildBench.

\end{abstract}

\section{Introduction}




Large language models (LLMs) such as GPT-4 \citep{achiam2023gpt4} and Llama-3 \citep{llama3} have become integral to AI applications due to their exceptional performance on a wide array of tasks by following instructions. The success of LLMs is heavily reliant on the data used for instruction fine-tuning, which equips them to handle a diverse range of tasks, including those not encountered during training. The effectiveness of instruction tuning depends crucially on access to high-quality instruction datasets. However, the alignment datasets used for fine-tuning models like Llama-3-Instruct are typically private, even when the model weights are open, which impedes the democratization of AI and limits scientific research for understanding and enhancing LLM alignment.

To address the challenges in constructing high-quality instruction datasets, researchers have developed two main approaches. The first type of method involves human effort to generate and curate instruction data \citep{Dolly, openassist, zhao2024wildchat, zheng2024lmsyschatm, zheng2024judging}, which is both \emph{time-consuming} and \emph{labor-intensive} \citep{liu2024best}. In contrast, the second type of methods uses LLMs to produce synthetic instructions~\citep{ding2023ultrachat, yin2023dynosaur, li2024synthetic, sun2024principle,alpaca, wang-etal-2023-self-instruct, wang2024codeclm, xu2023wizardlm, xu-etal-2023-baize, li2023camel}. Although these methods reduce human effort, its success heavily depends on prompt engineering and the careful selection of initial seed questions. The \emph{diversity} of synthetic data tends to decrease as the dataset size grows. Despite ongoing efforts, the scalable creation of high-quality and diverse instruction datasets continues to be a challenging problem.

\textit{Is it possible to synthesize high-quality instructions at scale by directly extracting data from advanced aligned LLMs?} A typical input to an aligned LLM contains three key components: the pre-query template, the query, and the post-query template. For instance, an input to Llama-2-chat could be ``\texttt{\small [INST]} Hi! \texttt{\small [/INST]}'', where \texttt{\small [INST]} is the pre-query template and \texttt{\small [/INST]} is the post-query template. These templates are predefined by the creators of the aligned LLMs to ensure the correct prompting of the models.
We observe that when we only input the pre-query template to aligned LLMs such as Llama-3-Instruct, they \textit{self-synthesize} a user query due to their auto-regressive nature. Our experiments indicate that these random user queries are of high quality and great diversity, suggesting that the abilities learned during the alignment process are effectively utilized.

\begin{figure}[t]
    \vspace{-1em}
    \includegraphics[width=1\textwidth]{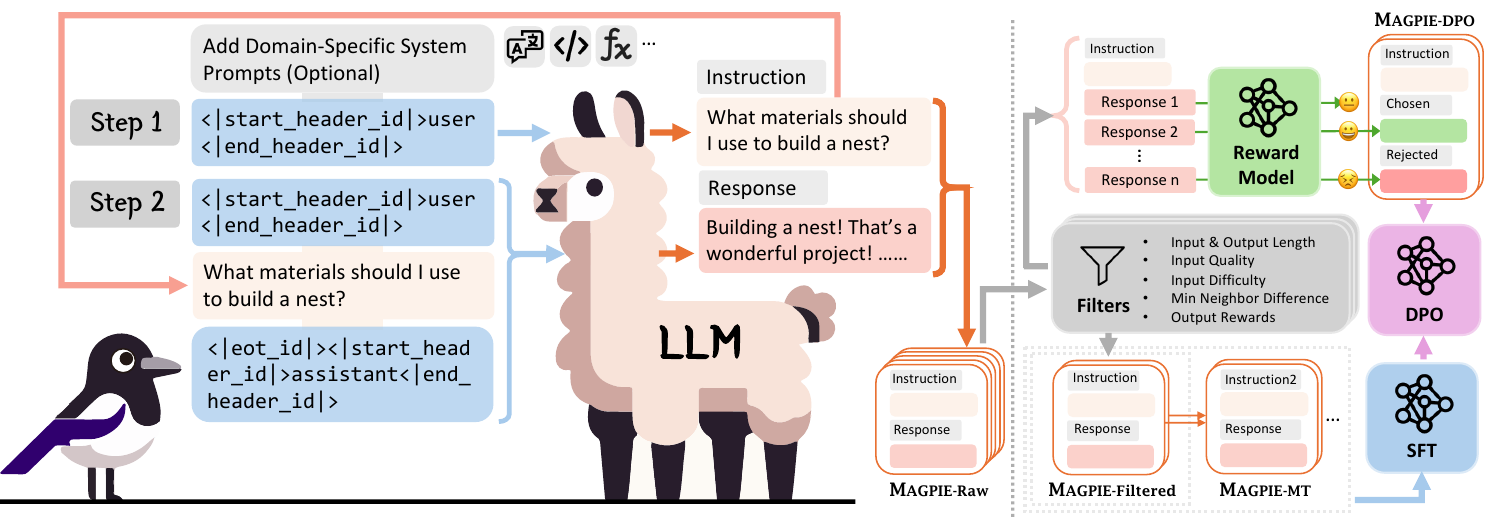}
    \caption{This figure illustrates \dataname{}, the process of self-synthesizing alignment data from aligned LLMs (e.g., Llama-3-8B-Instruct) to create a high-quality instruction dataset. 
    In Step 1, we input only the pre-query template into the aligned LLM and generate an instruction along with its response using auto-regressive generation. 
    In Step 2, we use a combination of a post-query template and another pre-query template to wrap the instruction generated from Step 1, prompting the LLM to generate the response. This completes the construction of the instruction dataset. \dataname{} efficiently generates diverse and high-quality instruction data, which can be further extended to multi-turn (\dataname{}-MT), preference optimization (\dataname{}-DPO), domain-specific, and multilingual datasets.
    }
    \vspace{-1em}
    \label{fig: main}
\end{figure}

Based on these findings, we developed a self-synthesis method to construct high-quality instruction datasets at scale, named \dataname{} (as illustrated in Figure \ref{fig: main}). Unlike existing methods, our approach does not rely on prompt engineering or seed questions. Instead, it \textit{directly} constructs instruction data by prompting aligned LLMs with a pre-query template for sampling instructions. We also demonstrated the extensibility of \dataname{} in generating multi-turn, preference optimization, domain-specific, and multilingual datasets.
We applied \dataname{} to the Llama-3-8B-Instruct and Llama-3-70B-Instruct models, creating two instruction datasets: \dataname{}-Air and \dataname{}-Pro, respectively.

Our \dataname{}-Air and \dataname{}-Pro datasets were created using 206 and 614 GPU hours, respectively, without any human intervention or API access to production LLMs like GPT-4. 
The statistics and advantages of \dataname{} datasets compared to existing ones are summarized in Table \ref{tab:compare with baselines} in Appendix \ref{appendix: statistics of magpie family}. We perform a comprehensive analysis of these two datasets in Section \ref{sec:analysis}, allowing practitioners to filter and select data instances for fine-tuning models according to their particular needs.

To compare \dataname{} data with other public instruction datasets (e.g., ShareGPT \citep{vicuna2023}, WildChat \citep{zhao2024wildchat}, Evol Instruct \citep{xu2023wizardlm}, UltraChat \citep{ding2023ultrachat}, OpenHermes \citep{OpenHermes, OpenHermes2.5}, GenQA \citep{chen2024genqa}, Tulu V2 Mix \citep{tulu2}), we conducted supervised fine-tuning (SFT) of the Llama-3-8B-Base model with each dataset and assess the performance of the fine-tuned models on alignment benchmarks such as AlpacaEval 2 \citep{alpaca_eval}, Arena-Hard \citep{arenahard2024}, and WildBench \citep{wildbench2024}.
Our results show that models supervised fine-tuned with \dataname{} achieve superior performance, even surpassing models that utilize both SFT and direct preference optimization (DPO) \citep{rafailov2024direct} with UltraFeedback \citep{cui2023ultrafeedback}. Notably, \dataname{}-aligned models outperform the official Llama-3-8B-Instruct model on AlpacaEval 2, despite the latter being fine-tuned with over 10 million data points for SFT and subsequent preference optimization.
Not only does \dataname{} excel in SFT alone compared to prior public datasets, but also delivers the best results when combined with preference optimization methods such as DPO. 
By leveraging \dataname{} extensions to generate high-quality preference optimization datasets, \dataname{}-aligned Llama-3 models
can even outperform GPT-4-Turbo(1106) on AlpacaEval 2.
These findings show the exceptional quality of instruction data generated by \dataname{}, enabling it to outperform even the official, extensively optimized, and proprietary LLMs.

\section{\dataname{}: A Scalable Method to Synthesize Alignment Data}\label{sec: method}

\color{black}
\textbf{Chat Templates of Aligned LLMs.} 
For an aligned LLM (e.g., Llama-3-8B-Instruct), the input sequence can be represented as $x = T_{pre-query} \oplus q \oplus T_{post-query}$. Here, $q$ is the user query (e.g., "What material should I use to build a nest?"), while $T_{pre-query}$ and $T_{post-query}$ are pre-query and post-query templates.
The pre-query template shows up before the user query, and the post-query template is defined as the conversation template between the user query and the LLM response. 
These templates are defined by the model provider to ensure the correct prompting. 
For example, for Llama-3-8B-Insturct model, $T_{pre-query}$ = \texttt{<|start\_header\_id|>user<|end\_header\_id|>}, and $T_{post-query}$ =\texttt{<|eot\_id|><|start\_header\_id|>assistant<|end\_header\_id|>}.

\color{black}

\subsection{\dataname{} Pipeline}
\label{sec: magpie pipeline}
\textbf{Overview of \dataname.}
In what follows, we describe our lightweight and effective method, \dataname, to synthesize alignment data from aligned LLMs. 
An instance of instruction data consists of at least one or multiple instruction-response pairs.
Each pair specifies the roles of instruction provider (e.g., user) and follower (e.g., assistant), along with their instruction and response.
As shown in Figure \ref{fig: main}, \dataname~consists of two steps: (1) instruction generation, and (2) response generation.
The \dataname{} pipeline can be fully \emph{automated without any human intervention}, and can be readily adapted for the generation of multi-turn, preference, and domain-specific datasets, as detailed in Section \ref{sec: magpie extension}.
We describe each step in the following.

\textbf{Step 1: Instruction Generation.}
The goal of this step is to generate an instruction for each instance of instruction data.
Given an open-weight aligned LLM (e.g., Llama-3-70B-Instruct), \dataname~crafts a pre-query template in the format of the predefined instruction template of the LLM.
Note that the auto-regressive LLM has been fine-tuned using instruction data in the format of the pre-query template.
Thus, the LLM autonomously generates an instruction when the pre-query template crafted by \dataname~is given as an input.
\dataname~stops generating the instruction once the LLM produces an end-of-sequence token.
Sending the crafted query to the LLM multiple times leads to a set of instructions.
We note that compared with existing synthetic approaches \citep{ding2023ultrachat, li2024synthetic, alpaca, wang-etal-2023-self-instruct, wang2024codeclm, xu2023wizardlm, xu-etal-2023-baize}, \dataname~does not require specific prompt engineering techniques since the crafted query follows the format of the predefined instruction template.
In addition, \dataname~autonomously generates instructions without using any seed question, ensuring the diversity of generated instructions.

\textbf{Step 2: Response Generation.} 
The goal of this step is to generate responses to the instructions obtained from Step 1.
\dataname~sends these instructions to the LLM to generate the corresponding responses.
Combining the roles of instruction provider and follower, the instructions from Step 1, and the responses generated in Step 2 yields the instruction dataset. 

\textbf{Applicability of \dataname{} on Different LLMs.} \dataname{} can be readily deployed to state-of-the-art open-weight language models including but not limited to Llama-3 \citep{llama3}, Llama-3.1 \citep{dubey2024llama}, Qwen2 \citep{yang2024qwen2}, Gemma-2 \citep{team2024gemma}, and Phi-3 \citep{abdin2024phi}. Please refer to Appendix \ref{appendix: statistics of magpie family} for detailed support and corresponding datasets.

\textcolor{black}{\textbf{Remark.} \dataname{} generates high-quality instructions even when the instruction loss is masked during alignment. We hypothesize that LLMs retain an implicit memorization of instruction distributions. We leave it as a potential future research problem.}

\subsection{\dataname{} Extensions}
\label{sec: magpie extension}

\textbf{Dataset Filtering.}
\dataname{} allows practitioners to select instruction data from the raw dataset generated from the above two steps based on their needs.
In Appendix \ref{appendix: filter setup}, we explores potential filter configurations with eight available metrics for users to customize their own \dataname{} datasets. We also provide 6 off-the-shelf filter configurations and discuss their performance in Appendix \ref{appendix: ablation on filter design}.

\textbf{Generating Multi-Turn Instruction Datasets.} \dataname{} can be readily extended to generate multi-turn instruction datasets. To construct a multi-turn dataset (denoted as \dataname-MT), we initially follow Steps 1 and 2 to generate the first turn of instruction and response. For subsequent turns, we append the pre-query template to the end of the full prompt from the previous round of communication. 
We observe that the model may occasionally forget its role as the user, especially for the 8B model. To mitigate this, we employ a system prompt designed to control the behavior of the LLM and reinforce its awareness of the multi-round conversation context. 
The full prompt for building the instructions of \dataname-MT can be found in Figure \ref{fig: generating mt prompt} in Appendix \ref{appendix: prompt template}. We follow the procedure in Step 2 of Section \ref{sec: magpie pipeline} to generate responses to form the multi-turn instruction dataset.

\textbf{Generating Preference Optimization Datasets.} 
Leveraging the diverse and high-quality instructions produced by \dataname{}, we present a simple and effective method for generating preference optimization data, inspired by \citet{meng2024simpo} and \citet{viethoangtranduong}. 
We first select a small proportion of high-quality instructions from the raw dataset generated by the \dataname{} pipeline, ensuring diverse task categories. For each selected instruction, we sample responses from the aligned LLM $k$ times, using a temperature of $T < 1$. We then employ a reward model (RM) to annotate scores for these responses. The response with the highest RM score is labeled as the chosen response, while the one with the lowest RM score is designated as the rejected response. 

\begin{figure}[t]
    \centering
    \vspace{-2em}
    \includegraphics[width=1\textwidth]{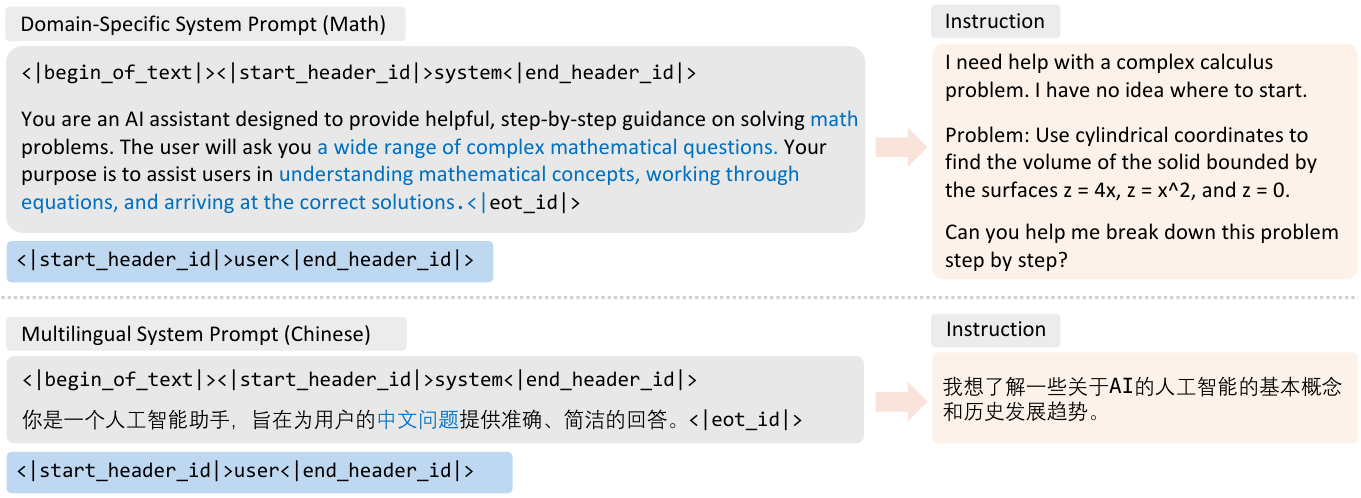}
    \caption{This figure illustrates how to control generating math and Chinese instructions using system prompts. The system prompt in Chinese translates to: "You are an AI assistant designed to provide accurate and concise answers to users' questions in Chinese." The user's response in Chinese translates to: "I want to learn about some basic concepts and historical development trends of AI."
    }
    \label{fig: system control topic}
\end{figure}

\textbf{Generating Domain-Specific and Multilingual Datasets.} In certain scenarios, users may wish to fine-tune LLMs using domain-specific or multilingual instruction data to enhance performance within specific domains or languages. To address this need, we introduce a lightweight method to control both the task category and the language of generated instructions. Our approach involves guiding LLMs through a tailored system prompt, specifying that the model is a chatbot designed for a particular domain and outlining the types of user queries it might encounter. Figure \ref{fig: system control topic} shows how to control generating math and Chinese instructions using system prompts.
More examples of system prompts designed to control the generation of code, translation, and multilingual instructions are illustrated in Figure \ref{fig: control generation topic} in Appendix \ref{appendix: prompt template}.

Furthermore, we note that domain-specific and multilingual instruction data can also be generated using models that are tailored to particular fields. \dataname{} demonstrates broad applicability beyond diverse chat models, extending to specialized code models (e.g., DeepSeek-Coder-V2 \citep{zhu2024deepseek}) and math models (e.g., Qwen2-Math-7B-Instruct \citep{yang2024qwen2}). By leveraging the unique strengths and specializations of different models, \dataname{} can create a rich and diverse corpus of instructions. Examples of \dataname{}-generated instructions from different domain-specific models and multilingual models are provided in Appendix \ref{appendix: magpie examples}.

\color{black}

\section{Dataset Analysis}\label{sec:analysis}


To demonstrate the effectiveness of \dataname{} compared with baseline methods for generating diverse high-quality alignment datasets, we apply \dataname{} to the Llama-3-8B-Instruct and Llama-3-70B-Instruct models \citep{llama3} to construct two instruction datasets: Llama-3-\dataname{}-Air (hereafter referred to as \dataname{}-Air) and Llama-3-\dataname{}-Pro (hereafter referred to as \dataname{}-Pro), respectively.
Examples of instances in both datasets can be found in Appendix \ref{appendix: magpie examples}.
In this section, we present a comprehensive analysis of the \dataname{}-Air and \dataname{}-Pro datasets, including topic coverage, difficulty, quality, similarity of instructions, and the quality of the responses. We also provide the safety analysis and cost analysis of \dataname{}.



\subsection{Dataset Coverage}

We follow \citet{zhao2024wildchat} and analyze the coverage of \dataname{}-Pro in the embedding space.
Specifically, we use the \texttt{all-mpnet-base-v2} embedding model\footnote{\url{https://huggingface.co/sentence-transformers/all-mpnet-base-v2}} to calculate the input embeddings, and employ t-SNE \citep{van2008visualizing} to project these embeddings into a two-dimensional space. We adopt three synthetic datasets as baselines, including \textbf{Alpaca} \citep{alpaca}, \textbf{Evol Instruct} \citep{xu2023wizardlm}, and \textbf{UltraChat} \citep{ding2023ultrachat}, to demonstrate the coverage of \dataname{}-Pro. The detailed analysis can be found in Appendix \ref{appendix: additional data coverage analysis}. We observe that the t-SNE plot of \dataname{}-Pro encompasses the area covered by the plots of Alpaca, Evol Instruct, and UltraChat. This suggests that \dataname{}-Pro provides a broader or more diverse range of topics. 

We also follow \citet{wang-etal-2023-self-instruct} and present the most common verbs and their top direct noun objects in instructions in Appendix \ref{appendix: more data analysis}, indicating the diverse topic coverage of \dataname{} dataset. 

\subsection{Dataset Attributes}

\paragraph{Attribute: Task Categories of Instructions.}
We use Llama-3-8B-Instruct to categorize the instances in \dataname{}-Pro (see Figure \ref{fig: llm task category stats} in Appendix \ref{appendix: additional data coverage analysis} for detail). 
The prompts used to query Llama-3-8B-Instruct can be found in Appendix \ref{appendix: prompt template}.
Our observations indicate that over half of the tasks in \dataname{}-Pro pertain to information seeking, making it the predominant category. 
This is followed by tasks involving creative writing, advice seeking, planning, and math. 
This distribution over the task categories aligns with the practical requests from human users \citep{alpaca_eval}.
\begin{wrapfigure}{r}{0.35\textwidth}
  \centering
  \includegraphics[width=0.35\textwidth]{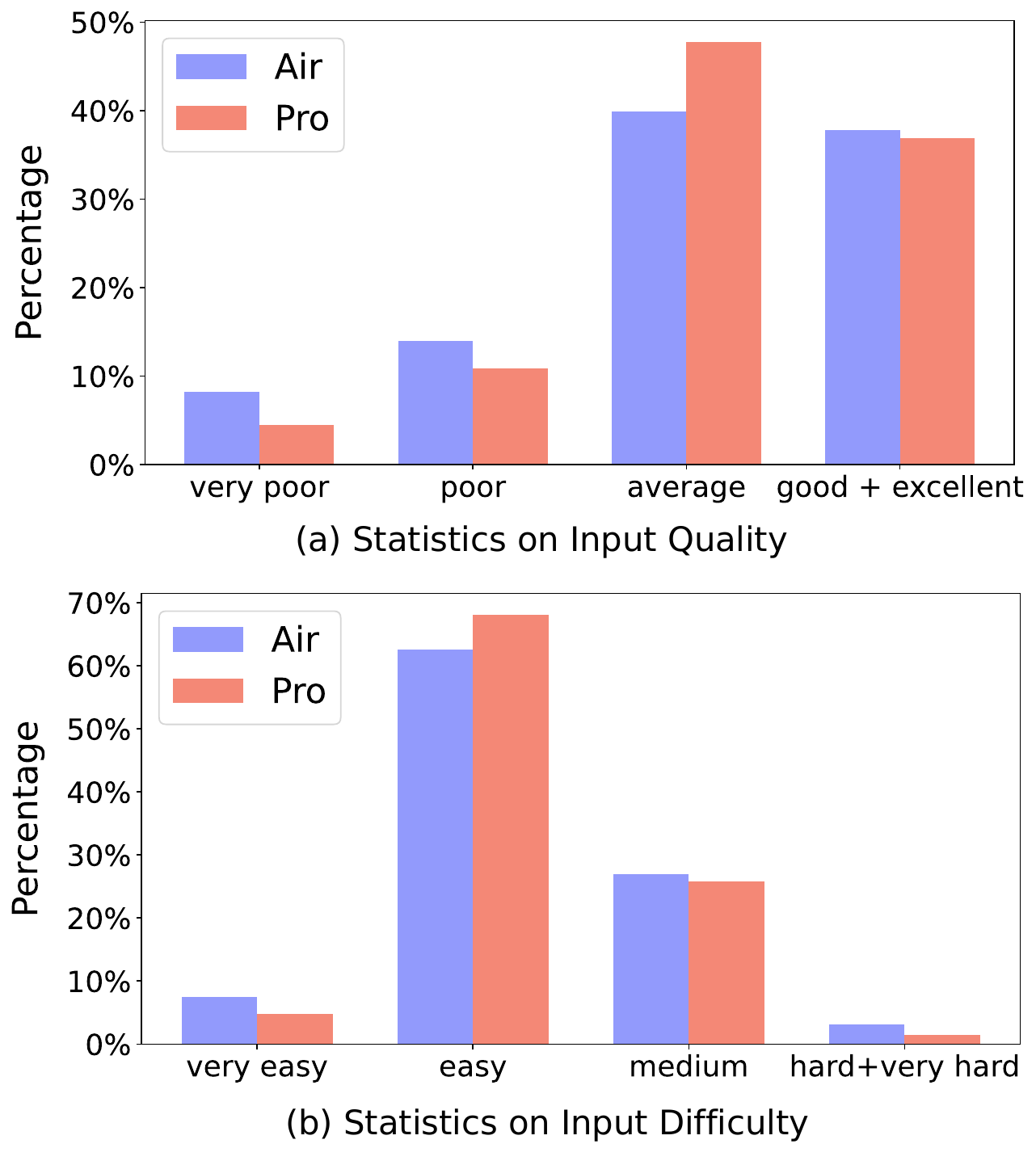}
  \caption{The statistics of instruction difficulty and quality.}
  \label{fig: llm input difficulty and quality stats}
  \vspace{-2em}
\end{wrapfigure}


\textbf{Attribute: Quality of Instructions.}
Similar to methods in \citep{chen2023alpagasus}, we prompt the Llama-3-8B-Instruct model to assess the quality of each instruction in \dataname{}-Air and \dataname{}-Pro, categorizing them as `very poor', `poor', `average', `good', and `excellent'.
We present the histograms of qualities for both datasets in Figure \ref{fig: llm input difficulty and quality stats}-(a).
We have the following two observations. 
First, both datasets are of high quality, with the majority of instances rated `average' or higher.
In addition, the overall quality of \dataname{}-Pro surpasses that of \dataname{}-Air. 
We hypothesize that this is due to the enhanced capabilities of Llama-3-70B compared with Llama-3-8B. 

\textbf{Attribute: Difficulty of Instructions.}
We use the Llama-3-8B-Instruct model to rate the difficulty of each instruction in \dataname{}-Air and \dataname{}-Pro.
Each instruction can be labeled as `very easy', `easy', `medium', `hard', or `very hard'.
Figure \ref{fig: llm input difficulty and quality stats}-(b) presents the histograms of the levels of difficulty for \dataname{}-Air and \dataname{}-Pro.
We observe that the distributions across difficulty levels are similar for \dataname{}-Air and \dataname{}-Pro.
Some instructions in \dataname{}-Pro are more challenging than those in \dataname{}-Air because \dataname{}-Pro is generated by a more capable model (Llama-3-70B-Instruct).

\begin{wrapfigure}{r}{0.35\textwidth}
  \centering
  \includegraphics[width=0.35\textwidth]{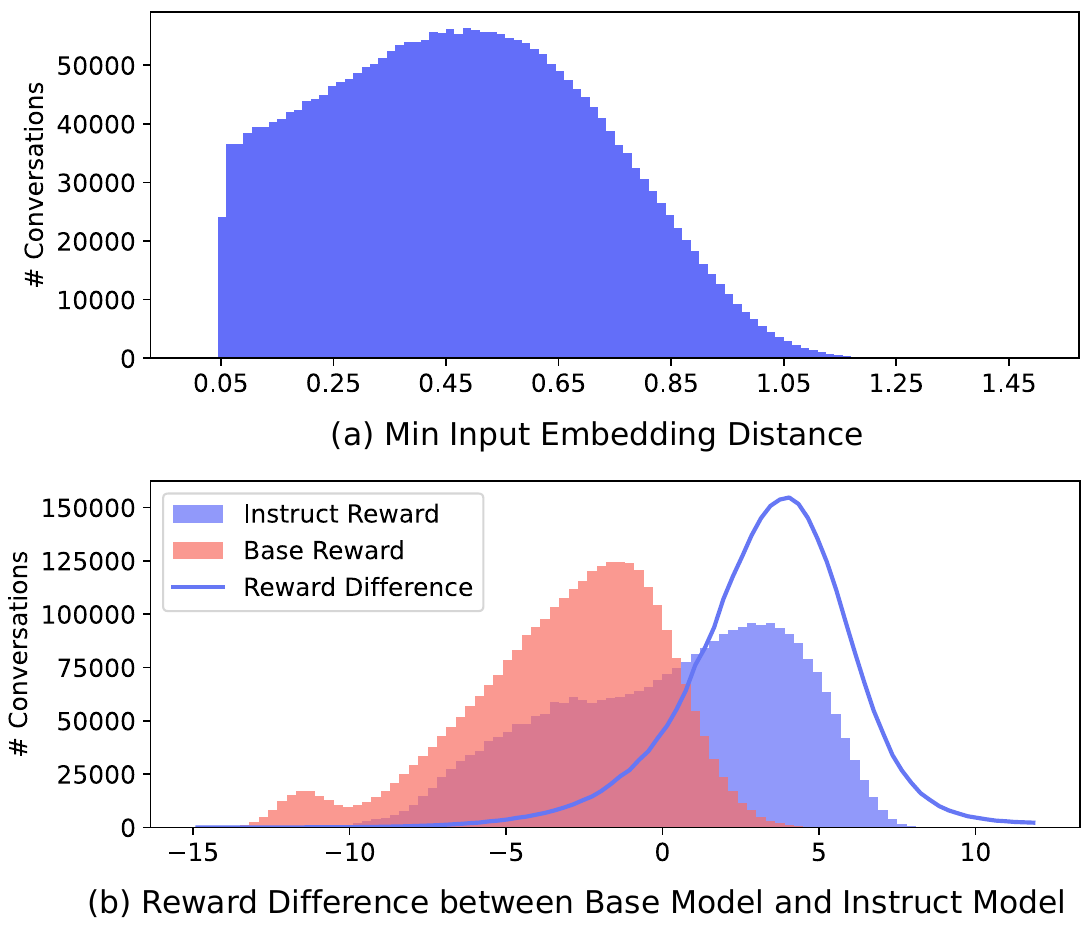}
  \caption{This figure summarizes the minimum neighbor distances and reward differences.}
  \label{fig: stats distance reward diff}
  \vspace{-1em}
\end{wrapfigure}

\textbf{Attribute: Instruction Similarity.}
We quantify the similarity among instructions generated by \dataname{} to remove repetitive instructions.
We measure the similarity using \textbf{minimum neighbor distance} in the embedding space.
Specifically, we first represent all instructions in the embedding space using the \texttt{all-mpnet-base-v2} embedding model.
For any given instruction, we then calculate the minimum distance from the instruction to its nearest neighbors in the embedding space using Facebook AI Similarity Search (FAISS) \citep{douze2024faiss}.
The minimum neighbor distances of instructions in \dataname{}-Air after removing repetitions are summarized in Figure \ref{fig: stats distance reward diff}-(a).

\textbf{Attribute: Quality of Responses.}
We assess the quality of responses using \textbf{rewards} assigned by a reward model, denoted as $r^*$.
For each instance in our dataset, we also calculate \textbf{reward difference} as $r^*-r_{base}$, where $r_{base}$ is the reward assigned by the same reward model to the response generated by the Llama-3 base model for the same instruction. We use URIAL \citep{lin2023unlocking} to elicit responses from the base model.
A positive reward difference indicates that the response from our dataset is of higher quality, and could potentially benefit instruction tuning.
In our experiments, we follow \citet{lambert2024rewardbench} and choose \texttt{FsfairX-LLaMA3-RM-v0.1} \citep{xiong2024iterative} as the reward model. 
Our results on the reward difference are presented in Figure \ref{fig: stats distance reward diff}-(b).

\subsection{Safety Analysis} \label{section: safety}

We use Llama-Guard-2 \citep{metallamaguard2} to analyze the safety of \dataname{}-Air and \dataname{}-Pro. Our results indicate that both datasets are predominantly safe, with less than 1\% of the data potentially containing harmful instructions or responses. Please see Appendix \ref{appendix: additional safety analysis} for detailed safety analysis.

\subsection{Cost Analysis} \label{section: cost analysis}

We perform experiments on a server with four NVIDIA A100-SXM4-80GB GPUs, an AMD EPYC 7763 64-Core Processor, and 512 GB of RAM, using the VLLM inference framework \citep{kwon2023efficient}.
The models are loaded in the \texttt{bfloat16} format.

When creating the 3M \dataname{}-Air dataset, our \dataname{} spent 1.55 and 50 hours to generate the instructions (Step 1) and responses (Step 2), respectively.
For the 1M \dataname{}-Pro dataset, \dataname{} used 3.5 and 150 hours to generate the instructions (Step 1) and responses (Step 2), respectively. 
Compared to existing approaches to create instruction datasets, the pipeline of \dataname{} is fully automated without any human intervention or API access to advanced commercial models such as GPT-4 \citep{achiam2023gpt4}.
Consequently, \dataname{} is cost-effective and scalable.
On average, implementing \dataname{} on a cloud server\footnote{\url{https://lambdalabs.com/service/gpu-cloud}} would incur costs of \textbf{\$0.12} and \textbf{\$1.1} per 1,000 data instances for \dataname{}-Air and \dataname{}-Pro, respectively.

\subsection{Additional Analysis} 
Additional dataset analysis, including the impact of generation configurations on the quality and difficulty of the generated instructions, is detailed in Appendix \ref{appendix: ablation generation config}. We provide ablation analysis on annotating models for assessing quality and difficulty in Appendix \ref{appendix: impact of annotating model}.

\section{Performance Analysis}
In this section, we evaluate the quality of \dataname{}-generated datasets by utilizing them to align base models including Llama-3 \citep{llama3}, Qwen1.5 \citep{Qwen}, and Qwen2 \citep{yang2024qwen2}. 

\subsection{Experimental Setups}
\label{section: experiments setups}
\textbf{Baselines for Supervised Fine-Tuning and Preference Optimization.} We compare the family of instruction datasets generated by \dataname{} with eight SOTA open-source instruction datasets: \textbf{ShareGPT} \citep{vicuna2023}, \textbf{WildChat} \citep{zhao2024wildchat}, \textbf{Evol Instruct} \citep{xu2023wizardlm}, \textbf{UltraChat} \citep{ding2023ultrachat}, \textbf{GenQA} \citep{chen2024genqa}, \textbf{OpenHermes 1} \citep{OpenHermes}, \textbf{OpenHermes 2.5} \citep{OpenHermes2.5}, and \textbf{Tulu V2 Mix} \citep{tulu2}. ShareGPT and WildChat are representative human-written datasets containing 112K and 652K high-quality multi-round conversations between humans and GPT, respectively. 
Evol Instruct, UltraChat, and GenQA are representative open-source synthetic datasets. 
Following \cite{meng2024simpo}, we use the 208K sanitized version of Ultrachat provided by HuggingFace\footnote{\url{https://huggingface.co/datasets/HuggingFaceH4/ultrachat\_200k}}. OpenHermes 1, OpenHermes 2.5, and Tulu V2 Mix are crowd-sourced datasets consisting of a mix of diverse open-source instruction datasets, with 243K, 1M, and 326K conversations, respectively. We also create an instruction dataset with 100K conversations using the Self-Instruct \citep{wang-etal-2023-self-instruct} and Llama-3-8B-Instruct model, denoted as \textbf{Self-Instruct (Llama-3)}.

We compare the models aligned using \dataname{} with preference optimization baselines using direct preference optimization (DPO) \citep{rafailov2024direct}. Specifically, we follow \cite{meng2024simpo} and use the models fine-tuned with the UltraChat dataset (for instruction tuning) and \textbf{Ultrafeedback} dataset (for preference optimization) \citep{cui2023ultrafeedback}.

\textbf{\dataname{} Setups.}
To demonstrate the quality of \dataname{}-generated instruction datasets for SFT, we select the first 300K \textbf{\dataname{}-Air} and \textbf{\dataname{}-Pro} raw datasets generated by Llama-3-8B-Instruct and Llama-3-70B-Instruct models, respectively. Apart from these raw datasets, we also applied the filters detailed in Appendix \ref{appendix: filter setup} and created two filtered datasets: \textbf{\dataname{}-Air-Filtered} and \textbf{\dataname{}-Pro-Filtered}, each contains 300K conversations. For preference optimization, we generate two additional datasets: \textbf{\dataname{}-Air-DPO} (generated by Llama-3-8B-Instruct) and \textbf{\dataname{}-Pro-DPO} (generated by Llama-3-70B-Instruct) with $k=5$ and $T=0.8$, each contains 100K conversations. We use \texttt{RLHFlow/ArmoRM-Llama3-8B-v0.1} \citep{ArmoRM} as the reward model. 

\textbf{Model Alignment Details.} For supervised fine-tuning, we follow \cite{touvron2023llama} and use a cosine learning rate schedule with an initial learning rate of $2 \times 10^{-5}$ when fine-tuning Llama-3, Qwen1.5 and Qwen2 base models. 
The maximum sequence length is 8192. For DPO, we use a cosine learning rate of $5 \times 10^{-7}$. The detailed parameters can be found in Appendix \ref{appendix: Experimental Setups for Instruction Tuning and Preference Tuning}.
We follow the official instruction templates of each model.

\color{black}

\textbf{Evaluation Benchmarks.} 
We evaluate the performance of the aligned models using two widely adopted instruction-following benchmarks: AlpacaEval 2 \citep{alpaca_eval} and Arena-Hard \citep{arenahard2024}. 
AlpacaEval 2 consists of 805 representative instructions chosen from real user interactions. 
Arena-Hard is an enhanced version of MT-Bench \citep{zheng2024judging}, containing 500 challenging user queries. 
Both benchmarks employ a GPT evaluator to assess responses generated by the model of interest and a baseline model.
Specifically, we use GPT-4-Turbo (1106) and Llama-3-8B-Instruct as baselines for AlpacaEval 2.
By default, Arena-Hard uses GPT-4 (0314) as its baseline model.

\textbf{Metrics.} 
We adopt two metrics to measure the capabilities of instruction-following of fine-tuned models.
The first metric is the \textbf{win rate (WR)}, which calculates the fraction of responses that are favored by the GPT evaluator.
This metric is applied in both benchmarks including AlpacaEval 2 and Arena-Hard.
The second metric is the \textbf{length-controlled win rate (LC)} \citep{dubois2024length}, a debiased version of WR. 
The GPT evaluator considers the lengths of responses generated by the baseline model and model under evaluation when computing LC.
By accounting for response length, LC reduces its impact on the win rate.
This metric is specifically applied to the AlpacaEval 2 benchmark \citep{alpaca_eval}.

\textbf{More Experimental Setups.} We provide more detailed descriptions of our experimental setups, including more model alignment details and benchmark decoding hyper-parameters in Appendix \ref{appendix: detailed experimental setups}.
 
\subsection{Experimental Results}
\label{sec: experimental results}


\begin{table}[]
    \centering
    \caption{This table compares the performance of models instruction-tuned on the Llama-3-8B base models using \dataname{}-generated datasets and baseline datasets. We observe that models aligned with our datasets significantly outperform those aligned with baseline datasets of the same order of magnitude in terms of data size. In addition, our fine-tuned models achieve comparable performance to the official aligned model, despite only undergoing SFT with a much smaller dataset. Numbers in \textbf{bold} indicate that \dataname{} outperforms the official Llama-3-8B-Instruct model.
    }   
    \resizebox{\textwidth}{!}{
    \begin{tabular}{c l   c   c c c c c c
 c c c c}\toprule
    \multicolumn{2}{c}{\multirow{3}[3]{*}{\makecell{\textbf{Alignment Setup} \\ (Base LLM = \textbf{Llama-3-8B}) }}}   & \multirow{3}[3]{*}{\makecell{\textbf{\#Convs}}} & \multicolumn{6}{c}{\textbf{AlpacaEval 2}} & \textbf{Arena-Hard} \\ 
    & & &  \multicolumn{3}{c}{GPT-4-Turbo (1106)} & \multicolumn{3}{c}{Llama-3-8B-Instruct} & \\ \cmidrule(lr){4-6} \cmidrule(lr){7-9}
   &  & & LC (\%) & WR (\%) & SD & LC (\%) & WR (\%) & SD & WR(\%) \\ \midrule
     SFT & +Self-Instruct (Llama-3) \citep{wang-etal-2023-self-instruct} & 100K & 7.21 & 5.18 & 0.7 & 17.86 & 12.73 & 1.05 & 4.0\\
     & +ShareGPT \citep{vicuna2023} & 112K & 9.73 & 7.2 & 0.81 & 27.26 & 18.32 & 1.18 & 6.5\\
     & +Evol Instruct \citep{xu2023wizardlm} & 143K & 8.52 & 6.25 & 0.76 & 20.16 & 14.98 & 1.1 & 5.1 \\
     & +OpenHermes 1 \citep{OpenHermes} & 243K & 9.94 & 6.27 & 0.73 & 29.19 & 17.92 & 1.16 & 4.4 \\
     & +Tulu V2 Mix \citep{tulu2} & 326K & 9.91 & 7.94 & 0.86  & 24.28 & 18.64 & 1.18 & 5.4 \\ 
     & +WildChat \citep{zhao2024wildchat} & 652K & 14.62 & 10.58 & 0.92  & 34.85 & 26.57 & 1.32 & 8.7 \\ 
     & +OpenHermes 2.5 \citep{OpenHermes2.5} & 1M & 12.89 & 9.74 & 0.91  & 32.68 & 25.01 & 1.30 & 8.2 \\ 
     & +GenQA \citep{chen2024genqa} & 6.47M & 9.05 & 7.11 & 0.82 & 21.90 & 16.09 & 1.12 & 3.0 \\ 
     & +\underline{UltraChat} \citep{ding2023ultrachat} {\small{\rotatebox[origin=c]{315}{\faArrowRight}}} & 208K & 8.29 & 5.44 & 0.71 & 23.95 & 15.12 & 1.11 & 3.6 \\
    + DPO & \text{    }~~~~~ +UltraFeedback(\citep{cui2023ultrafeedback}) & 64K & 18.36 & 17.33 & 1.14 & 44.42 & 42.36 & 1.46 & 14.8 \\
     \midrule\midrule
     SFT & \multicolumn{1}{l}{\textbf{\dataname}-Air-300K-Raw}  & 300K & 21.99 & 21.65 & 1.21 & 48.63 & 48.06 & 1.42 & 15.8\\
     & \multicolumn{1}{l}{\underline{\textbf{\dataname}-Air-300K-Filtered} {\small{\rotatebox[origin=c]{315}{\faArrowRight}}}} & 300K & 22.66 & \textbf{23.99} & 1.24 & 49.27 & \textbf{50.8} & 1.44 & 14.9 \\ 
     + DPO & \text{    }~~~~~ +\textbf{\dataname}-Air-DPO & 100K & \textbf{45.48} & \textbf{50.43} & 1.48 & \textbf{75.06} & \textbf{79.64} & 1.18 & \textbf{35.9} \\ \midrule
     SFT & \multicolumn{1}{l}{\textbf{\dataname}-Pro-300K-Raw}  & 300K & 21.65 & 22.19 & 1.2 & 49.65 & \textbf{50.84} & 1.42 & 15.9 \\
     & \multicolumn{1}{l}{\underline{\textbf{\dataname}-Pro-300K-Filtered} {\small{\rotatebox[origin=c]{315}{\faArrowRight}}}} & 300K & \textbf{25.08} & \textbf{29.47} & 1.35 & \textbf{52.12} & \textbf{53.43} & 1.44 & 18.9 \\
     + DPO & \text{    }~~~~~ +\textbf{\dataname}-Pro-DPO & 100K & \textbf{50.10} & \textbf{53.53} & 1.45 & \textbf{78.52} & \textbf{80.82} & 1.17 & \textbf{35.7} \\ \midrule \midrule
    \rowcolor{gray!20} 
     \multicolumn{2}{l}{\quad Llama-3-8B-Instruct (SFT+DPO) } & \multirow{1}{*}{$>$10M\tablefootnote{\url{https://huggingface.co/meta-llama/Meta-Llama-3-8B-Instruct}}}  & 22.92 & 22.57 & 1.26 & 50 & 50 & - & 20.6 \\
     \bottomrule
    \end{tabular}}
    \label{tab:main}
\end{table}

\paragraph{\dataname{} datasets outperform baselines with SFT only.} 

In Table \ref{tab:main}, we compare the performance of Llama-3 models fine-tuned with instruction datasets generated by \dataname{} against those supervised fine-tuned with baseline datasets.
Using the AlpacaEval 2 benchmark, we observe that both the LC and WR of our supervised fine-tuned models surpass all those models fine-tuned with baseline SFT datasets.
This indicates that the datasets generated by \dataname{} are of higher quality, leading to significantly enhanced instruction-following capabilities. 
A similar observation is made when using the Arena-Hard evaluation benchmark.
We highlight that the Llama-3 base models supervised fine-tuned with instruction datasets generated by \dataname{} outperform even those models that have undergone preference optimization (i.e., STF followed by DPO), which further emphasizes the high quality of data generated by \dataname{}.
\begin{wrapfigure}{r}{0.4\textwidth}
  \centering
  \includegraphics[width=0.4\textwidth]{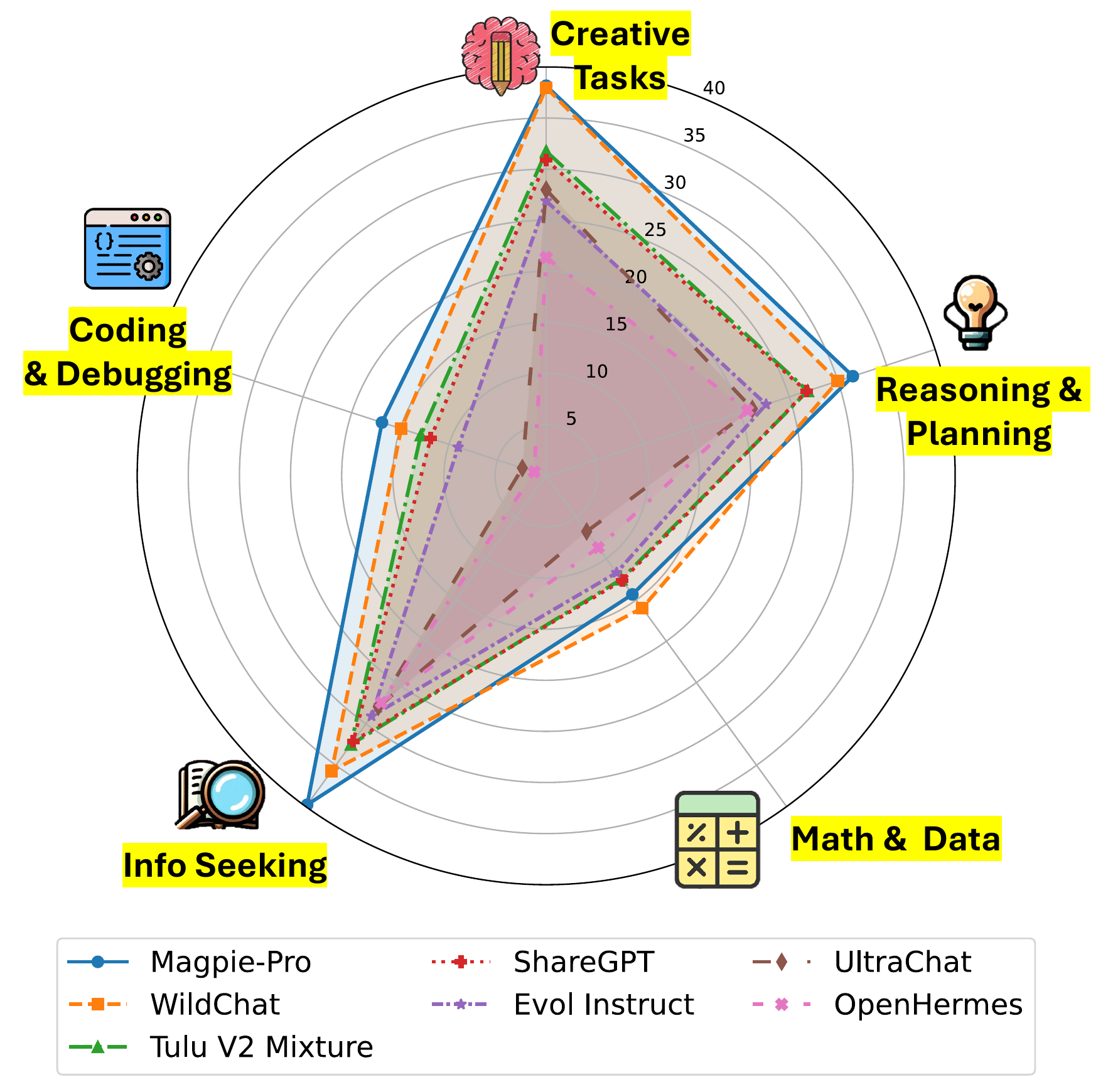}
  \caption{This figure shows the performance breakdown by category of \dataname{}-Pro and baselines on WildBench.}
  \label{fig: radar comparison}
  \vspace{-2em}
\end{wrapfigure}

\color{black}

To investigate the advantages of \dataname{} across different task categories, we also compare the performance of models fine-tuned with \dataname{}-Pro compared with baseline datasets using WildBench benchmark \citep{wildbench2024}. This benchmark consists of 1024 tasks carefully selected from real-world human-LLM conversation logs. The results are demonstrated in Figure \ref{fig: radar comparison}. We observe that \dataname{} consistently outperforms baseline datasets across categories.


\textbf{Models aligned with data generated by \dataname{} achieve comparable or even higher performance to the official aligned model, but with fewer data.} 
In Table \ref{tab:main}, we also compare the performance of models aligned with data generated by \dataname{} against the official aligned model (Llama-3-8B-Instruct).
We observe that the Llama-3-8B base model supervised fine-tuned with data from \dataname{} outperforms Llama-3-8B-instruct using the AlpacaEval 2 benchmark.
For example, when Llama-3-8B-Instruct is chosen as the baseline model of AlpacaEval 2, we observe that LC of Llama-3-8B base models fine-tuned with instruction data from \dataname{} exceeds 50\%, indicating a preference for our SFT models over the official aligned model. In addition, when DPO is applied, our aligned model demonstrates remarkable performance gains. Specifically, it outperforms the official Llama-3-8B-Instruct model on both the AlpacaEval 2 and Arena-Hard benchmarks. Most notably, our model even surpasses GPT-4-Turbo(1106) on AlpacaEval 2.
Finally, we highlight that our alignment process uses no more than 400K data, whereas the official aligned models are aligned with more than 10M data samples. 
This demonstrates the high quality of the data generated by \dataname{}.

\color{black}

\textbf{\dataname{} can enhance the performance of other backbone models.} 
Table \ref{tab: other backbone} illustrates the efficacy of \dataname{} when applied to generate instruction dataset and fine-tune other base models, i.e., Qwen2-1.5B, Qwen1.5-4B, and Qwen1.5-7B. The results demonstrate that our fine-tuned models achieve better performance than the official aligned models, which have undergone both supervised fine-tuning and preference tuning.
These results underscore the effectiveness of \dataname{} and the quality of its generated instructions. In addition, we apply \dataname{}-generated datasets to align Llama-3.1-Minitron-4B-Width-Base \citep{sreenivas2024llm} and Llama-3.1-8B-Instruct \citep{dubey2024llama} using SFT followed by DPO. The resulting aligned model, which we term MagpieLM, achieves remarkable performance and ranks first among popular open-source instruction models with fewer than 10 billion parameters. The details of MagpieLM are deferred to Appendix \ref{appendix: magpielm}.

    
    

\begin{table}[]
    \centering
    \caption{This table compares the performance of models instruction-tuned on the Qwen base models using the \dataname{}-Pro-300K-Filtered dataset and the official instruction-tuned models. The Qwen base model enhanced with \dataname{} outperforms the official instruction-tuned model.}
    \resizebox{0.8\textwidth}{!}{
    \begin{tabular}{c l c c c c c
 c c c c c c c }\toprule
    \multicolumn{2}{c}{\multirow{3}[3]{*}{\textbf{Alignment Setup}}} & \multicolumn{6}{c}{\textbf{AlpacaEval 2}}\\
    & &  \multicolumn{3}{c}{GPT-4-Turbo (1106)} & \multicolumn{3}{c}{Official Aligned Model as Ref.}\\ \cmidrule(lr){3-5} \cmidrule(lr){6-8}
    & &  LC (\%) & WR (\%) & SD & LC (\%) & WR (\%) & SD  \\ \midrule
    \multirow{2}{*}{Qwen2-1.5B} & Qwen2-1.5B-Instruct & \textbf{3.91} & 3.00 & 0.54 & 50 & 50 & -  \\
   & Base Model + \dataname & {3.48} & \textbf{5.32} & 0.67 & \textbf{56.66} & \textbf{66.27} & 1.50 \\ \midrule
   \multirow{2}{*}{Qwen1.5-4B} & Qwen1.5-4B-Chat & 5.89 & 4.74 & 0.67 & 50 & 50 & -  \\
   & Base Model + \dataname & \textbf{9.1} & \textbf{10.96} & 0.93 & \textbf{68.09} & \textbf{72.42} & 1.42 \\
   \midrule
   \multirow{2}{*}{Qwen1.5-7B} & Qwen1.5-7B-Chat & 14.75 & 11.77 & 0.97 & 50 & 50 & - \\
   & Base Model + \dataname & \textbf{15.10} & \textbf{18.51} & 1.14 & 46.28 & \textbf{58.53} & 1.44 \\
    \bottomrule
    \end{tabular}
    }
    \label{tab: other backbone}
    \vspace{-1em}
\end{table}

\textbf{Performance of \dataname{} on More Benchmarks.}
We report the performance of models supervised fine-tuned using \dataname{}-Air and \dataname{}-Pro, evaluated across a range of tasks featured on the Huggingface Open LLM Leaderboard \citep{open-llm-leaderboard} in Table \ref{tab: more benchmarks}. The tasks includes MMLU \citep{hendrycks2020measuring}, ARC Challenge \citep{clark2018think}, HellaSwag \citep{zellers2019hellaswag}, TruthfulQA \citep{lin2021truthfulqa}, WinoGrande \citep{levesque2012winograd}, and GSM8K \citep{cobbe2021training}. 
We also perform experiments on MMLU-Redux \citep{gema2024we} with zero-shot prompting. Our experimental results demonstrate that models fine-tuned with \dataname{}-Air and \dataname{}-Pro achieve comparable performance to the official instruct model and other baselines.

\color{black}
We note that the performance of \dataname{} may degrade on reasoning tasks, which is attributed to the small proportion of reasoning instructions in \dataname{}-Air and \dataname{}-Pro datasets.
In response, we provide a supplementary "booster" dataset containing 150K math, code, and reasoning instructions using the \dataname{} extension mentioned in Section \ref{sec: magpie extension}. We combine this booster dataset with \dataname{}-Pro-300K-Filtered and create \dataname{}-Pro-Mix-Filtered. Experimental results presented in Table \ref{tab: more benchmarks} demonstrate that the model supervised fine-tuned using the mixed dataset effectively addresses the initial weakness in reasoning tasks. Notably, this new model ranks among the top-3 of all model checkpoints, performing only slightly weaker than OpenHermes 2.5 (1M conversations) and Llama-3-8B-Instruct (>10M conversations). This significant improvement showcases the flexibility and adaptability of the \dataname{} framework in generating task-specific instruction data.

\begin{table}[]
    \centering
    \caption{This table compares the performance of models supervised-fine-tuned on \dataname{}-Air, \dataname{}-Pro, and \dataname{}-Pro-Mix against baselines and official instruct model across various downstream benchmarks. All models are supervised-fine-tuned on the Llama-8B base models.
    }
    \resizebox{\textwidth}{!}{
    \begin{tabular}{c| c c c c c c | c | c} \toprule
        \textbf{Alignment Setup} & \textbf{MMLU (5)} & \textbf{ARC (25)} & \textbf{HellaSwag (10)} & \textbf{TruthfulQA (0)} & \textbf{WinoGrande (5)} & \textbf{GSM8K (5)} & \textbf{MMLU-Redux (0)} & \textbf{Average} \\ \midrule
        ShareGPT & 66.03 & 58.45 & 81.50 & 52.34 & 74.03 & 48.67 & 50.68 & 61.67 \\
        Evol Instruct & 65.62 & 60.75 & 82.70 & 52.87 & 76.16 & 42.91 & 52.73 & 61.96 \\
        GenQA & 63.45 & 58.53 & 79.65 & 48.85 & 74.03 & 43.14 & 51.87 & 59.93 \\
        OpenHermes 1 & 65.42 & 62.29 & 82.15 & 50.85 & 75.61 & 47.16 & 46.07 & 61.36 \\
        OpenHermes 2.5 & 65.70 & 61.86 & 82.53 & 51.35 & 76.09 & 67.02 & 46.07 & \textbf{66.24} \\
        Tulu V2 Mix & 66.34 & 59.22 & 82.80 & 47.99 & 76.16 & 58.07 & 46.97 & 62.51 \\
        WildChat & 65.95 & 59.22 & 81.39 & 53.18 & 75.30 & 48.75 & 52.59 & 62.34 \\
        UltraChat & 65.23 & 62.12 & 81.68 & 52.76 & 75.53 & 50.57 & 50.75 & 62.66\\
        \midrule 
        \dataname{}-Air-300K-Filtered & 64.45 & 61.01 & 79.90 & 53.48 & 72.38 & 52.24 & 52.34 & 62.25 \\
        \dataname{}-Pro-300K-Filtered & 64.25 & 60.41 & 80.52 & 52.46 & 73.32 & 47.92 & 52.16 & 61.58\\
        \textbf{\dataname{}-Pro-Mix-Filtered} & 65.65 & 59.64 & 80.72 & 50.81 & 73.24 & 63.08 & 56.34 & \textbf{64.21} \\
        \midrule \midrule
        \rowcolor{gray!20} 
        Llama-3-8B-Instruct & 67.82 & 61.52 & 78.67 & 52.47 & 72.14 & 71.72 & 58.60 & \textbf{66.13} \\ 
        \bottomrule
    \end{tabular}
    }
    \vspace{-1em}
    \label{tab: more benchmarks}
\end{table}

\color{black}

\textbf{Additional Experimental Results.} We defer additional experimental results and analysis of multi-turn datasets, i.e., \dataname{}-Air-MT and \dataname{}-Pro-MT, to Appendix \ref{appendix: multi-round magpie}. We conduct a detailed comparison between \dataname{} and Self-Instruct in Appendix \ref{appendix: compare with self-instruct}. In addition, ablations on data quantity, quality, filter designs, and response generator are deferred in Appendices \ref{appendix: data quantity and quality}, \ref{appendix: ablation on filter design}, and \ref{appendix: ablation on output generator}.
\dataname{} model's performance on trustworthiness benchmarks is reported in Appendices \ref{appendix: compare with official}.
\section{Related Work}

\textbf{LLM Alignment.}
Instruction tuning \citep{wei2022finetuned} and preference tuning \citep{bai2022training} are widely used to align the responses of LLMs with human values.
Instruction tuning utilizes an instruction dataset to fine-tune LLMs, where each instruction data consists of one turn or multiple turns of instructions and desired responses. 
The performance of instruction tuning heavily relies on the quality of instruction data \citep{alpaca,wang-etal-2023-self-instruct,zhou2024lima}.
Preference tuning further improves responses of LLMs using reinforcement learning human feedback (RLHF) \citep{bai2022training} or preference optimization \citep{azar2024general, ethayarajh2024kto,hong2024reference,rafailov2024direct} based on a preference dataset.

\textbf{Alignment Dataset Construction.} We classify the existing methods of creating datasets for model alignment into two main categories: human interactions with LLMs and synthetic instruction generation.
To create datasets for alignment, previous studies have collected \textbf{human} interactions with LLMs \citep{Dolly, zhao2024wildchat, zheng2024lmsyschatm, zheng2024judging, openassist}. However, manually crafting instructions is not only time-consuming and labor-intensive, but may also incorporate toxic content \citep{zhao2024wildchat}. 
Another category of approaches \citep{wang-etal-2023-self-instruct, alpaca, xu2023wizardlm, xu-etal-2023-baize, wang2024codeclm, sun2024principle} focus on prompting LLMs to generate \textbf{synthetic} instruction datasets, beginning with a small set of human-annotated seed instructions and expanding these through few-shot prompting. However, these methods face a diversity challenge, as few-shot prompting often results in new instructions that are too similar to the original seed questions \citep{li2024synthetic}. To enhance coverage, some research \citep{ding2023ultrachat, li2024synthetic} summarizes world knowledge and employs it to generate synthetic datasets. We note that our \dataname{} dataset also belongs to the synthetic dataset. However, we leverage the prompt template without any requirement for seed questions or prompt engineering.

Compared to the above two main categories, alignment data can also be generated by \textbf{transforming} existing data~\citep{wang-etal-2022-super, sanh2022multitask, gandhi2024better}. However, the constrained variety of NLP tasks in these datasets may impede the ability of tuned LLMs to generalize in real-world scenarios \citep{li2024synthetic}. There are also \textbf{mixture} datasets (e.g., \citep{tulu2, OpenHermes, liu2024deita, zhou2024lima}) that combine or select high-quality instruction data from various existing open-source instruction datasets to enhance coverage \citep{tulu2, OpenHermes} and/or improve overall performance \citep{liu2024deita, zhou2024lima}. 
There are also data construction methods focusing on improving the reasoning and math abilities~\citep{Yue2023MAmmoTHBM, yue2024mammoth2}, which can be further merged with \dataname{} for creating a better mixture of data for instruction tuning.

\textbf{Training Data Extraction.} Language models have the capability to memorize examples from their training datasets, potentially enabling malicious users to extract private information \citep{brown2022does, biderman2024emergent, carlini2021extracting}. Pioneering work \citep{Krishna2020Thieves, carlini2021extracting, nasr2023scalable} has demonstrated that it is possible to extract private pre-training data from BERT \citep{devlin2018bert}, GPT-2 \citep{radford2018improving}, and ChatGPT \citep{achiam2023gpt4}, respectively.  \cite{yu2023bag} propose several techniques including adjusting sampling strategies to better extract training datasets from language models.
Recently, \cite{kassem2024alpaca} propose a black-box prompt optimization method that uses an attacker LLM to extract high levels of memorization in a victim LLM. \cite{wang2024pandora} leverage membership inference attack (MIA) to extract fine-tuning datasets from fine-tuned language models. \cite{bai2024special} extract the training dataset of production language models via special characters (e.g., structural symbols of JSON files, and \@, \# in emails and online posts).
Different from the prior work, we aim to create publicly available alignment datasets with minimal human effort by leveraging the remarkable generation capabilities of LLMs, rather than extracting private training data from LLMs.
\section{Limitations, Discussions, and Ethical Considerations}
\label{section: limitations}

\textbf{Limitations and Discussions.} 
\dataname{}-aligned LLMs demonstrate strong performance on instruction following benchmarks compared to the official Llama-3-8B-Instruct. However, we observe a performance degradation on math and reasoning benchmarks. Although we leverage the techniques described in Section \ref{sec: magpie extension} to generate specialized booster reasoning datasets, there is still a performance gap between \dataname{}-aligned LLMs and the official models. Enhancing the reasoning ability of \dataname{}-aligned models presents a promising direction for future research.

\color{black}



\textbf{Societal Impact and Potential Harmful Consequences.}
The primary objective of this paper is to develop a scalable method to synthesize instruction data to enhance the instruction-following capabilities of LLMs, and thus align them with human values.
However, the data generated by \dataname{} may contain harmful instructions and/or responses, which may lead to unsafe behaviors if used raw in instruction tuning. 
Our empirical evaluations indicate that such harmful data instances constitute less than 1\% of the dataset.
Our data filtering technique in Appendix \ref{appendix: filter setup} can identify and remove these instances, thus mitigating the risk.

\section{Conclusion}

In this paper, we developed a scalable method, \dataname, to synthesize instruction data for fine-tuning large language models. 
\dataname{} leveraged the predefined instruction templates of open-weight LLMs and crafted a prompt specifying only the role of instruction provider.
Given the crafted prompt, the LLM then generated detailed instructions due to their auto-regressive nature.
\dataname{} then sent the generated instructions to the LLM to generate corresponding responses.
These pairs of instructions and responses constituted the instruction dataset.
We used Llama-3-8B-instruct to label the instruction dataset and developed a filtering technique to select effective data instances for instruction tuning.
We fine-tuned the Llama-3-8B base model using the selected data, and demonstrated that the fine-tuned model outperformed those fine-tuned using all baselines.
Moreover, our fine-tuned models outperformed the official aligned model, Llama-3-8B-Instruct, which has been instruction-tuned and preference-optimized using more than 10M data instances.
This highlighted the quality of the instruction data synthesized by \dataname.

\section{Acknowledgement}

The research of Z. Xu, F. Jiang, L. Niu, and R. Poovendran is partially supported by the National Science Foundation (NSF) AI Institute for Agent-based Cyber Threat Intelligence and Operation (ACTION) under grant IIS 2229876 and Air Force Office of Scientific Research (AFOSR) under grant FA9550-23-1-0208.
The research of Y. Choi is partially supported by the National Science Foundation (NSF) under grant DMS-2134012 (Scaling Laws of Deep Learning) and the Office of Naval Research (ONR) under grant N00014-24-1-2207 (Symbolic Knowledge Distillation of LLMs for All: Diverse Scales, Skills, and Values).

This work is supported in part by funds provided by the National Science Foundation, Department of Homeland Security, and IBM. 
Any opinions, findings, and conclusions or recommendations expressed in this material are those of the author(s) and do not necessarily reflect the views of the NSF or its federal agency and industry partners.

\bibliography{iclr2025_conference}
\bibliographystyle{iclr2025_conference}

\appendix
\clearpage
\color{black}
\section{Statistics of instruction datasets generated by \dataname{} compared to other instruction datasets.}
\label{appendix: statistics of magpie family}

\dataname{} can be readily deployed to state-of-the-art open-weight model families, including but not limited to Llama-3 \citep{llama3}, Llama-3.1 \citep{dubey2024llama}, Qwen2 \citep{yang2024qwen2}, Gemma-2 \citep{team2024gemma}, and Phi-3 \citep{abdin2024phi} model families.

In what follows, we compare datasets generated by \dataname{} with the above model families compared to other state-of-the-art instruction datasets.
The \dataname{} dataset family encompasses over 11.4 million diverse and high-quality instructions and corresponding responses generated from state-of-the-art open-source models. This corpus represents the largest alignment dataset for LLMs that does not rely on human-written questions or employ complex multi-stage pipelines.

\begin{table}[htbp]
\label{table:dataset_stats} 
\caption{Statistics of \dataname{} family compared to other instruction datasets. Tokens are counted using the \texttt{tiktoken} library \citep{tiktoken}.
}
\resizebox{\textwidth}{!}{
\begin{tabular}{@{}c lccccccc@{}}
\toprule
\makecell{Instruction\\Source} & Dataset Name & \#Convs & \#Turns &\makecell{Human\\Effort} & \makecell{Response\\Generator} & \#Tokens / Turn & \#Total Tokens \\ \midrule
\multirow{3}{*}{Synthetic} & Alpaca \citep{alpaca} & 52K & 1 & Low & text-davinci-003 & 67.38$_{\pm{54.88}}$ & 3.5M \\
& Evol Instruct \citep{xu2023wizardlm} & 143K & 1 & Low & ChatGPT & 473.33$_{\pm{330.13}}$ & 68M  \\
& UltraChat \citep{ding2023ultrachat} & 208K & 3.16 & Low & GhatGPT & 376.58$_{\pm{177.81}}$ & 238M \\
\midrule
\multirow{4}{*}{Human}  & Dolly \citep{Dolly} & 15K & 1 & High & ChatGPT & 94.61$_{\pm{135.84}}$ & 1.42M  \\
& ShareGPT \citep{zheng2024judging} & 112K & 4.79 & High & ChatGPT & 465.38$_{\pm{368.37}}$ & 201M  \\
& WildChat \citep{zhao2024wildchat} & 652K & 2.52 & High & GPT-3.5 \& GPT-4 & 727.09$_{\pm{818.84}}$ & 852M \\
& LMSYS-Chat-1M \citep{zheng2024lmsyschatm}& 1M & 2.01 & High & Mix & 260.37$_{\pm{346.97}}$ & 496M \\
\midrule
\multirow{3}{*}{Mixture} & Deita \citep{liu2024deita}& 9.5K & 22.02 & - & Mix & 372.78$_{\pm{182.97}}$ & 74M \\
& OpenHermes \citep{OpenHermes} & 243K & 1 & - & Mix & 297.86$_{\pm{258.45}}$ & 72M \\
& Tulu V2 Mixture \citep{tulu2} & 326K & 2.31 & - & Mix & 411.94$_{\pm{447.48}}$ & 285M \\
\midrule \midrule
\multirow{10}{*}{\bf \dataname{}} & Llama-3-\dataname{}-Air & \textbf{3M} & 1 & \bf No & Llama-3-8B-Instruct & 426.39$_{\pm{217.39}}$ & 1.28B\\
& Llama-3-\dataname{}-Air-MT & 300K & 2 & \bf No & Llama-3-8B-Instruct & 610.80$_{\pm{90.61}}$ & 366M \\
& Llama-3-\dataname{}-Pro & 1M & 1 & \bf No & Llama-3-70B-Instruct & 478.00$_{\pm{211.09}}$ & \textcolor{black}{477M} \\
& Llama-3-\dataname{}-Pro-MT & 300K & 2 & \bf No & Llama-3-70B-Instruct & 554.53$_{\pm{133.64}}$ & 333M \\

& Llama-3.1-\dataname{}-Pro & 1M & 1 & \bf No & Llama-3.1-70B-Instruct & 482.35$_{\pm{378.45}}$ & \textcolor{black}{482M} \\
& Llama-3.1-\dataname{}-Pro-MT & 300K & 2 & \bf No & Llama-3.1-70B-Instruct & 552.53$_{\pm{325.49}}$ & 331M \\

& Qwen2-\dataname{}-Air & 3M & 1 & \bf No & Qwen2-7B-Instruct & 577.87$_{\pm{416.10}}$ & \textcolor{black}{1.73B} \\
& Qwen2-\dataname{}-Pro & 1M & 1 & \bf No & Qwen2-72B-Instruct & 424.87$_{\pm{339.71}}$ & 424M \\

& Gemmma-2-\dataname{}-Pro & 1M & 1 & \bf No & Gemma-2-27b-it & 483.90$_{\pm{237.80}}$ & \textcolor{black}{259M} \\
& Phi-3-\dataname{}-Pro & 534K & 1 & \bf No & Phi-3-Medium-Instruct & 391.38$_{\pm{414.32}}$ & 391M \\

\bottomrule
\end{tabular}
}
\label{tab:compare with baselines}
\end{table}

\color{black}

\section{MagpieLM}
\label{appendix: magpielm}

\begin{figure}[htbp]
    \centering
    \includegraphics[width=0.8\textwidth]{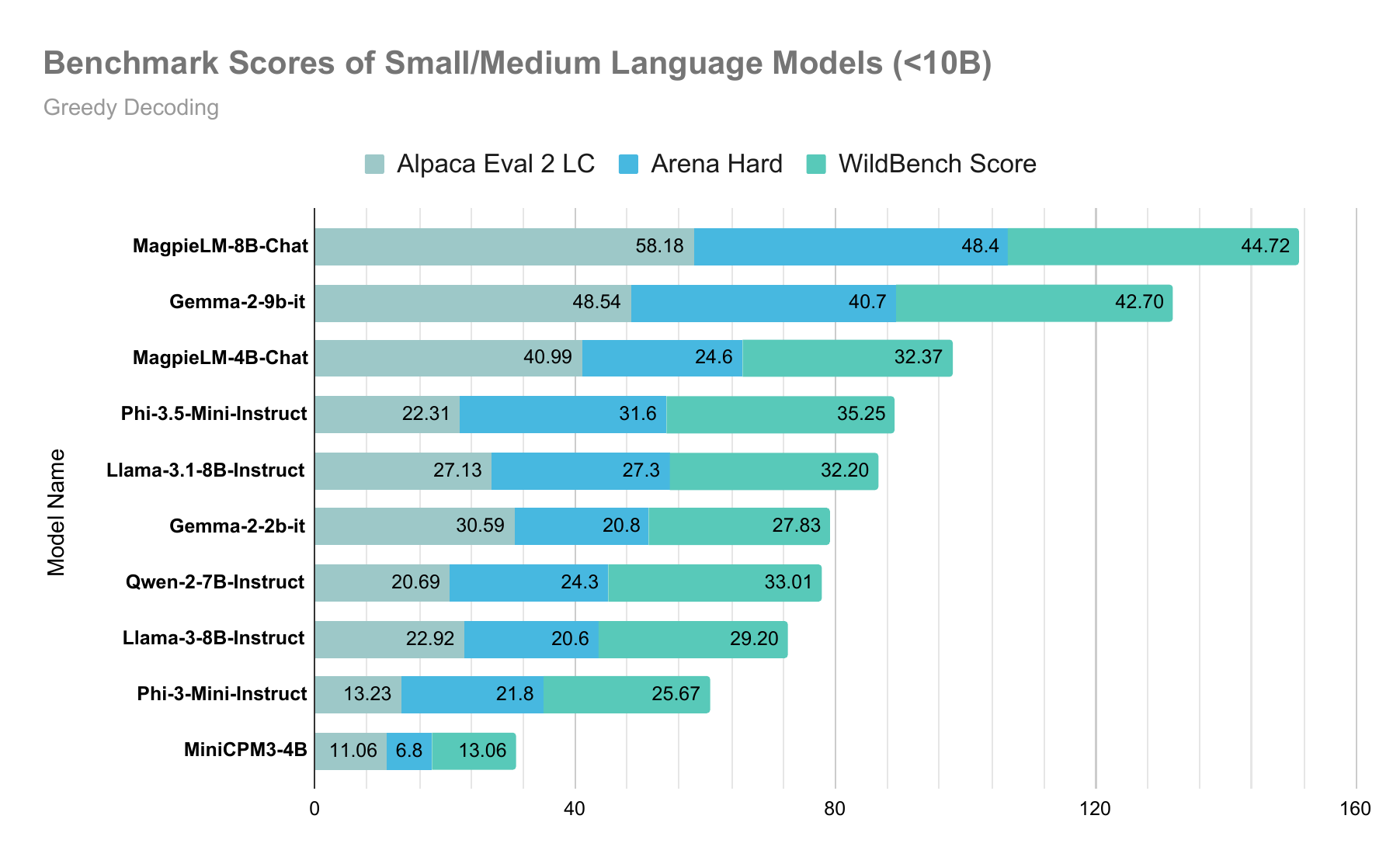}
    \caption{This figure shows the performance of \textsc{MagpieLM}-4B-Chat and \textsc{MagpieLM}-8B-Chat compared with baselines. \textsc{MagpieLM} significantly outperforms baselines of similar sizes.}
    \label{fig: magpielm_bench}
\end{figure}

In this section, we discuss the details of our \textsc{MagpieLM}. To construct the SFT and DPO datasets for aligning \textsc{MagpieLM}, we select 550K and 200K high-quality instructions, respectively, from the \dataname{} family. These instructions cover diverse categories, ensuring a comprehensive training set. We then generate corresponding responses using the Gemma-2-9b-it model \citep{team2024gemma}.

The benchmark performance of both models is demonstrated in Figure \ref{fig: magpielm_bench}. Notably, \textsc{MagpieLM} significantly outperforms other baselines of similar model sizes across multiple benchmarks, including Alpaca Eval 2 \citep{alpaca_eval}, Arena Hard \citep{arenahard2024}, and Wildbench \citep{wildbench2024}.

\section{Filter Setups}\label{appendix: filter setup}

In this section, we explore potential filter configurations for selecting high-quality instructional data for fine-tuning purposes. We provide the following metrics to enable users to customize their filtered \dataname{} dataset:

\begin{enumerate}
\item \textbf{Input Length}: The total number of characters in the instructions.
\item \textbf{Output Length}: The total number of characters in the responses.
\item \textbf{Task Category}: The specific category of the instructions. See Appendix \ref{appendix: additional data coverage analysis} for details.
\item \textbf{Input Quality}: The clarity, specificity, and coherence of the instructions, rated as `very poor', `poor', `average', `good', and `excellent'.
\item \textbf{Input Difficulty}: The level of knowledge required to address the task described in the instruction, rated as `very easy', `easy', `medium', `hard', or `very hard'.
\item \textbf{Minimum Neighbor Distance}: The embedding distance to the nearest neighbor. Can be used for filtering out repetitive or similar instances.
\item \textbf{Reward}: Denoted as $r^*$. See Section \ref{sec:analysis} for details. This metric can be used to filter out low-quality responses, such as repetitions or refusals.
\item \textbf{Reward Difference}: Denoted as $r^*-r_{base}$. See Section \ref{sec:analysis} for details.
\end{enumerate}

We provide several off-the-shelf configurations, as demonstrated in Table \ref{tab: filter design}. We defer the detailed performance analysis of each filter configuration for \dataname{}-Pro to Appendix \ref{appendix: ablation on filter design}.

\begin{table}[htbp]
\caption{Different filter configurations we provide. We note that the Output Length filter is applied last. Specifically, this filter selects the $k$ instances of the longest responses. In our experiments, we empirically set $\tau_1=-12$, and $\tau_2=0$.}
\vspace{1em}
\resizebox{\textwidth}{!}{
\begin{tabular}{@{}ccccccccccc@{}}
\toprule
\makecell{Source Dataset} & Filter Name & \#Convs & \makecell{Input\\Length} & \makecell{Output\\Length} & \makecell{Task\\Category} & \makecell{Input\\Quality} & \makecell{Input\\Difficulty} & \makecell{Min Neighbor\\Distance} & Reward & \makecell{Reward\\Difference}\\ \midrule

\multirow{1}{*}{\dataname{}-Air} & Filter & 300K & - & Longest & - & $\geq$ good & $\geq$ medium & $> 0$ & - & $>\tau_2$ \\ \midrule
\multirow{5}{*}{\dataname{}-Pro} & Filter & 300K & - & Longest & - & $\geq$ average & - & $> 0$ & $>\tau_1$ & - \\
& Filter2 & 300K & - & Longest & - & $\geq$ good & $\geq$ easy & $> 0$ & $>\tau_1$ & - \\
& Filter3 & 300K & - & Longest & - & - & - & $> 0$ & $>\tau_1$ & - \\
& Filter4 & 300K & - & Longest & - & $\geq$ good & $\geq$ easy & $> 0$ & - & $>\tau_2$ \\
& Filter5 & 338K & - & - & - & $\geq$ good & $\geq$ easy & $> 0$ & $>\tau_1$ & - \\
& Filter6 & 200K & - & Longest & - & - & $50\%$ easy + $50\%$ $>$ easy & $> 0$ & $>\tau_1$ & - \\
\bottomrule
\end{tabular}
}

\label{tab: filter design}
\end{table}

\section{More Dataset Analysis}
\label{appendix: more data analysis}
This section provides additional dataset analysis, complementing the discussions in Section \ref{sec:analysis}. Statistics including lengths of instructions and responses are illustrated in Figure \ref{fig: input_output_distance stats}.

\subsection{Additional Analysis on Dataset Coverage and Attributes.}
\label{appendix: additional data coverage analysis}

\begin{figure}[htbp]
\vspace{-1em}
    \centering
    \begin{minipage}{0.36\textwidth}
    \centering
    \begin{subfigure}[b]{1\textwidth}
        \centering
        \includegraphics[width=\textwidth]{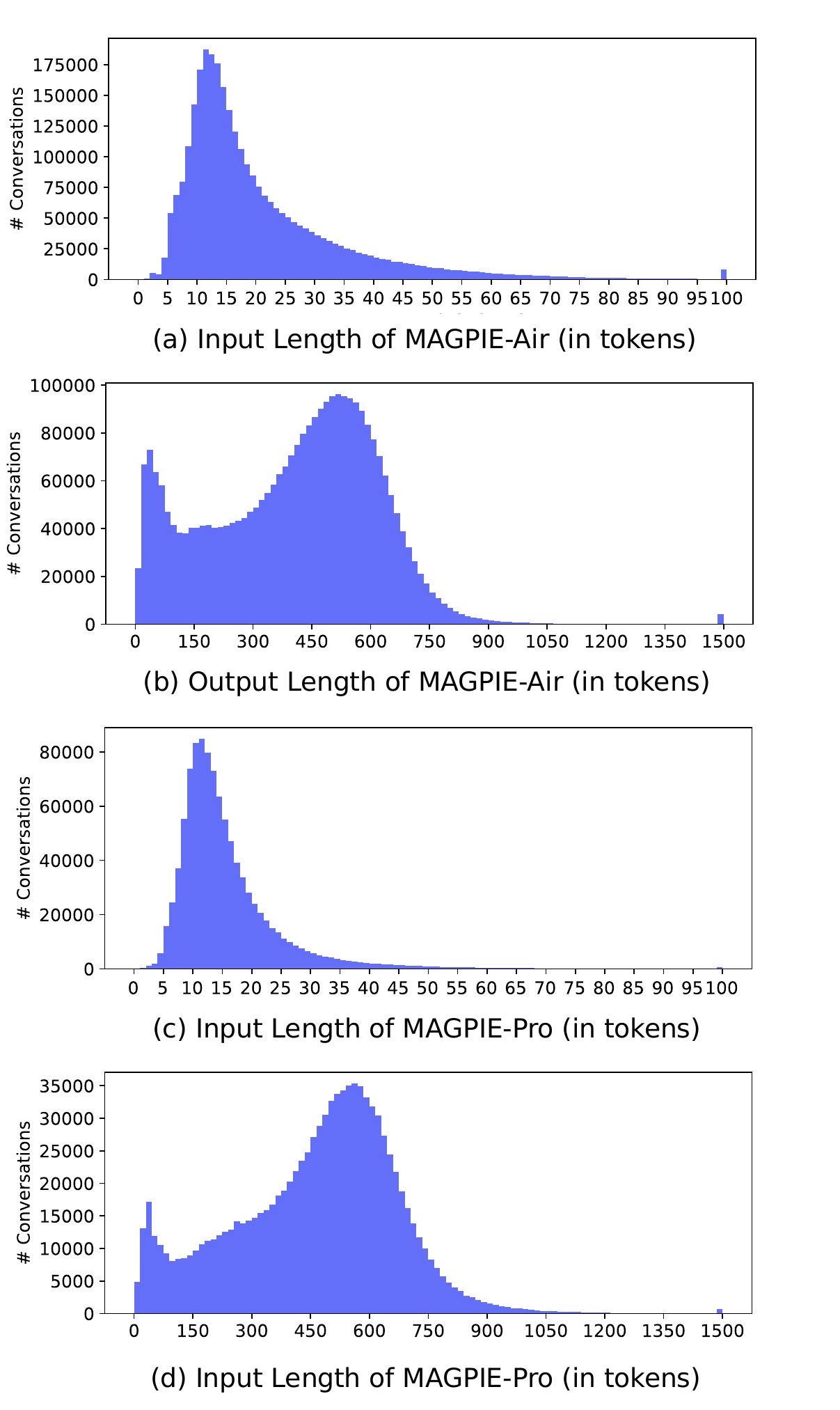}
    \end{subfigure}
    \vspace{-1em}
    \caption{Lengths of instructions and responses in \dataname{}-Air/Pro.}
    \vspace{-1em}
    \label{fig: input_output_distance stats}
    \end{minipage}
    ~
    \begin{minipage}{0.62\textwidth}
    \centering
    \vspace{-2em}
    \includegraphics[width=1\textwidth]{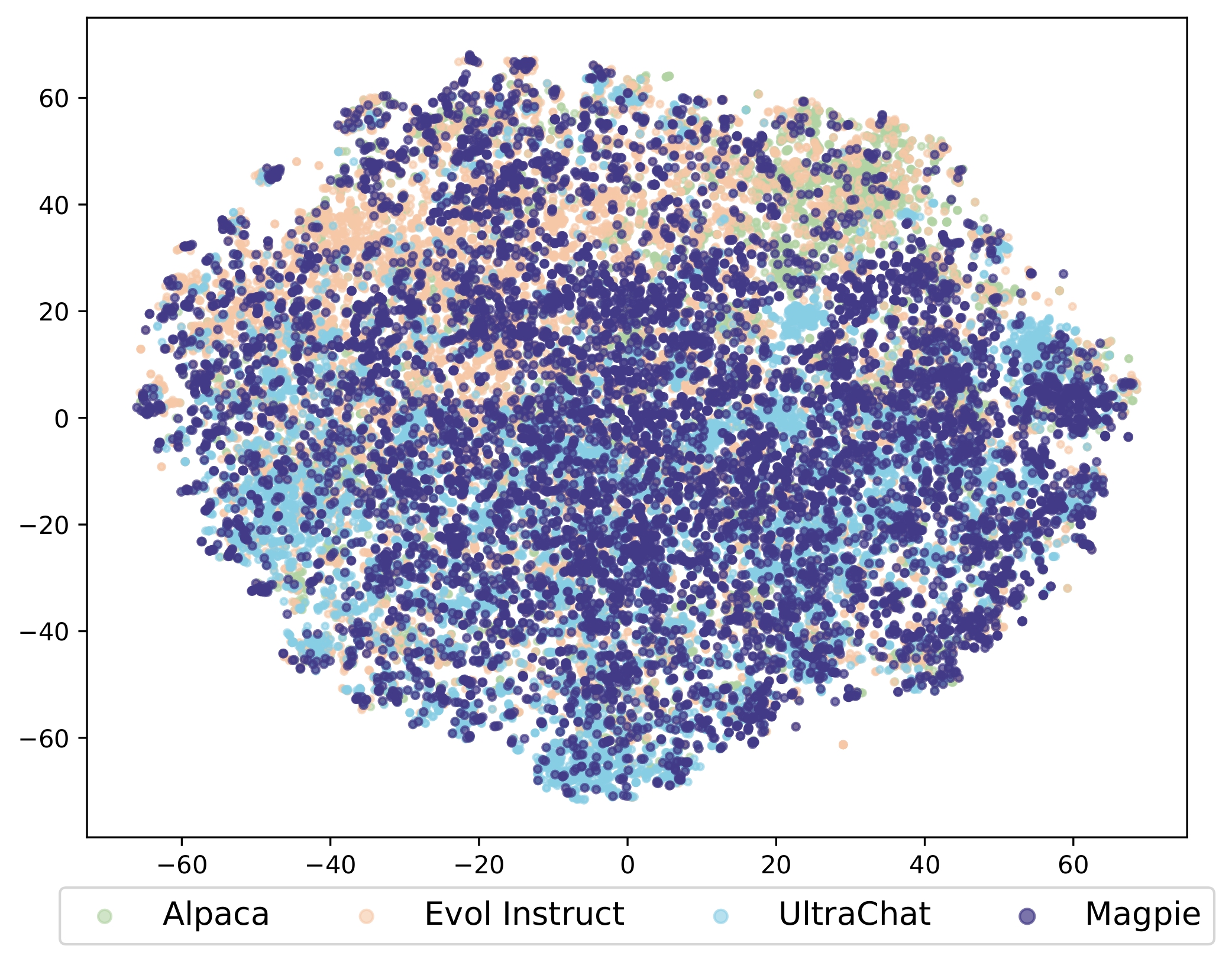}
    \vspace{-1em}
    \caption{This figure compares the t-SNE plot of \dataname{}-Pro with those of Alpaca, Evol Instruct, and UltraChat, each of which is sampled with 10,000 instructions. 
    The t-SNE plot of \dataname{}-Pro encompasses the area covered by the other plots, demonstrating the comprehensive coverage of \dataname{}-Pro. 
    \vspace{-3em}
    }
    \label{fig: tsne-synthetic}
    \end{minipage}
\end{figure}

\paragraph{Dataset Coverage Measured by T-SNE.} Figure \ref{fig: tsne-synthetic} presents the t-SNE plots of \dataname{}-Pro, Alpaca, Evol Instruct, and UltraChat. Each t-SNE plot is generated by randomly sampling 10,000 instructions from the associated dataset.
We observe that the t-SNE plot of \dataname{}-Pro encompasses the area covered by the plots of Alpaca, Evol Instruct, and UltraChat. 
This suggests that \dataname{}-Pro provides a broader or more diverse range of topics, highlighting its extensive coverage across varied themes and subjects. 

\begin{figure}[h]
    \centering
    \begin{subfigure}{0.43\textwidth}
    \centering
    \includegraphics[width=1\textwidth]{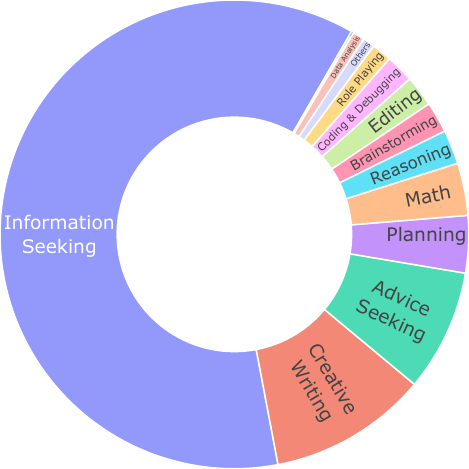}
    \vspace{-1em}
    \caption{Task categories of \dataname{}-Pro.
    }
    \label{fig: task-category-70b}
    \end{subfigure}
    \hspace{2em}
    \begin{subfigure}{0.43\textwidth}
    \centering
    \includegraphics[width=1\textwidth]{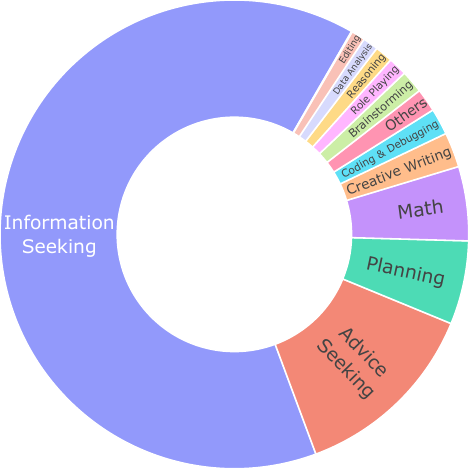}
    \vspace{-1em}
    \caption{Task categories of \dataname{}-Air.
    }
    \end{subfigure}
    \caption{This figure visualizes the task category of \dataname{}-Pro and \dataname{}-Air by topic tags.}
    \label{fig: llm task category stats}
\end{figure}

\paragraph{Task Categories of \dataname{}-Pro and \dataname{}-Air.}
Figure \ref{fig: llm task category stats} illustrates the task category distributions for \dataname{}-Pro and \dataname{}-Air, as labeled by Llama-3-Instruct. We observe that the task category distributions of these two datasets are largely similar, however, \dataname{}-Pro exhibits a higher percentage of creative writing tasks.


\begin{figure}[t]
\vspace{-1em}
    \centering
    \includegraphics[width=0.7\textwidth]{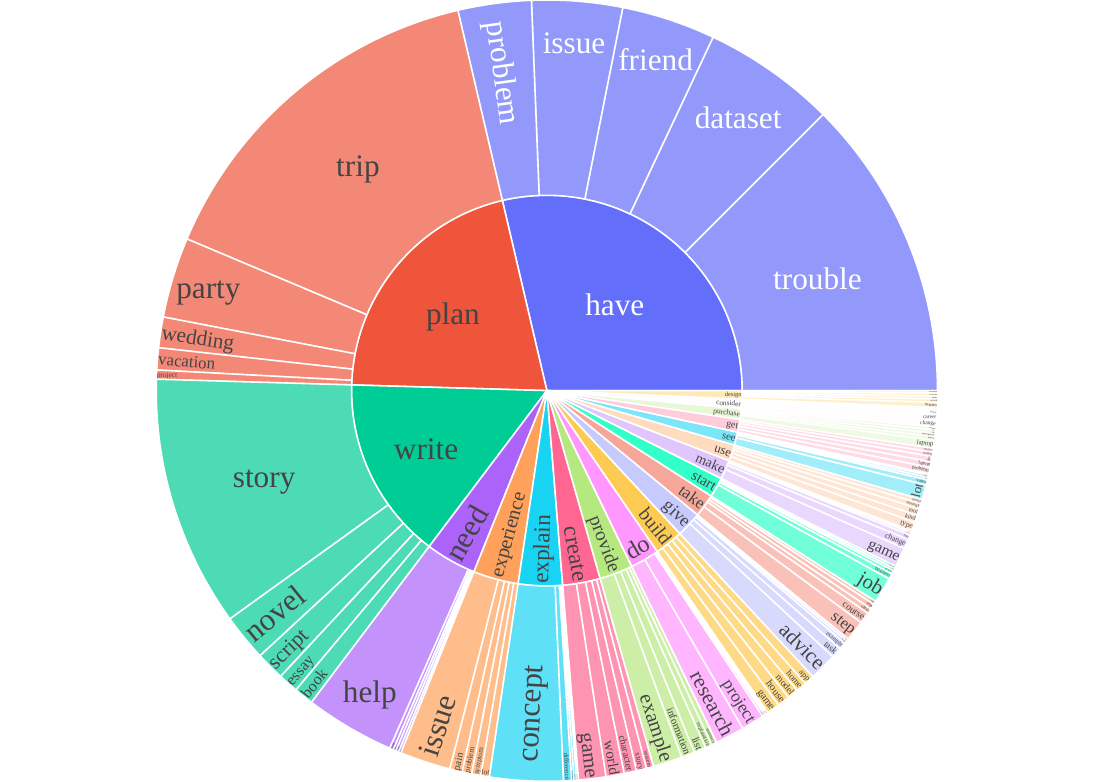}
    \vspace{0.5em}
    \caption{This figure demonstrates the top 20 most common root verbs (shown in the inner circle) and their top 5 direct noun objects (shown in the outer circle) within the \dataname{}-Air dataset. This indicates that \dataname{} encompasses a broad range of topics.}
    \label{fig: verb noun 8b}
\end{figure}

\paragraph{Visualization of Root Verbs and Their Direct Noun Objects.} 
Figure \ref{fig: verb noun 8b} visualizes the top common root verbs and their direct noun objects of \dataname{}-Air dataset. This indicates the diverse topic coverage of MAGPIE-Air.

\subsection{Additional Safety Analysis}
\label{appendix: additional safety analysis}

Table \ref{tab: safety analysis} illustrates the percentage of different unsafe categories of \dataname{}-Air and \dataname{}-Pro, as labeled by Llama-Guard-2 \citep{metallamaguard2}. We have two key observations. First, the proportion of data containing potentially harmful queries is minimal, with less than 1\% for both datasets. Second, the majority of unsafe responses fall into the category of specialized advice, which includes responses that may offer specialized financial, medical, or legal advice, or suggest that dangerous activities or objects are safe. 

\begin{table}[htbp]
    \centering
    \caption{This table shows the percentage of different unsafe categories of \dataname{}-Air and \dataname{}-Pro tagged by Llama-Guard-2 \cite{metallamaguard2} model.}
    \vspace{1em}
    \resizebox{\textwidth}{!}{
    \begin{tabular}{c c c c c c  c c c c c c c c } \toprule
        Dataset & Safe & \makecell{\small Violent \\ \small Crimes}& \makecell{Non-Violent\\ Crimes }& \makecell{Sex-Related \\ Crimes} & \makecell{Child Sexual \\ Exploitation} & \makecell{Specialized\\ Advice} & \makecell{Privacy}& \makecell{Intellectual \\ Property} &\makecell{Indiscriminate \\ Weapons} & Hate & \makecell{Suicide \& \\  Self-Harm} &\makecell{Sexual\\Content} & Others  \\ \midrule
        \dataname{}-Air & 99.128\% & 
0.001\% & 
0.073\% & 
0.003\% & 
0.000\% & 
0.636\% & 
0.022\% & 
0.026\% & 
0.038\% & 
0.001\% & 
0.002\% & 
0.009\% & 
0.062\% \\

        \dataname{}-Pro & 99.347\% & 0.001\% & 0.049\% & 0.002\% & 0.000\% & 0.446\% & 0.015\% & 0.074\% & 0.014\% & 0.001\% & 0.004\% & 0.011\% & 0.036\%  \\ 
        \bottomrule
    \end{tabular}
    }
    \label{tab: safety analysis}
\end{table}


\subsection{Ablation Analysis on Generation Configurations}
\label{appendix: ablation generation config}

\begin{figure*}[htbp]
    \centering
    \begin{subfigure}[b]{0.3\textwidth}
        \includegraphics[width=\textwidth]{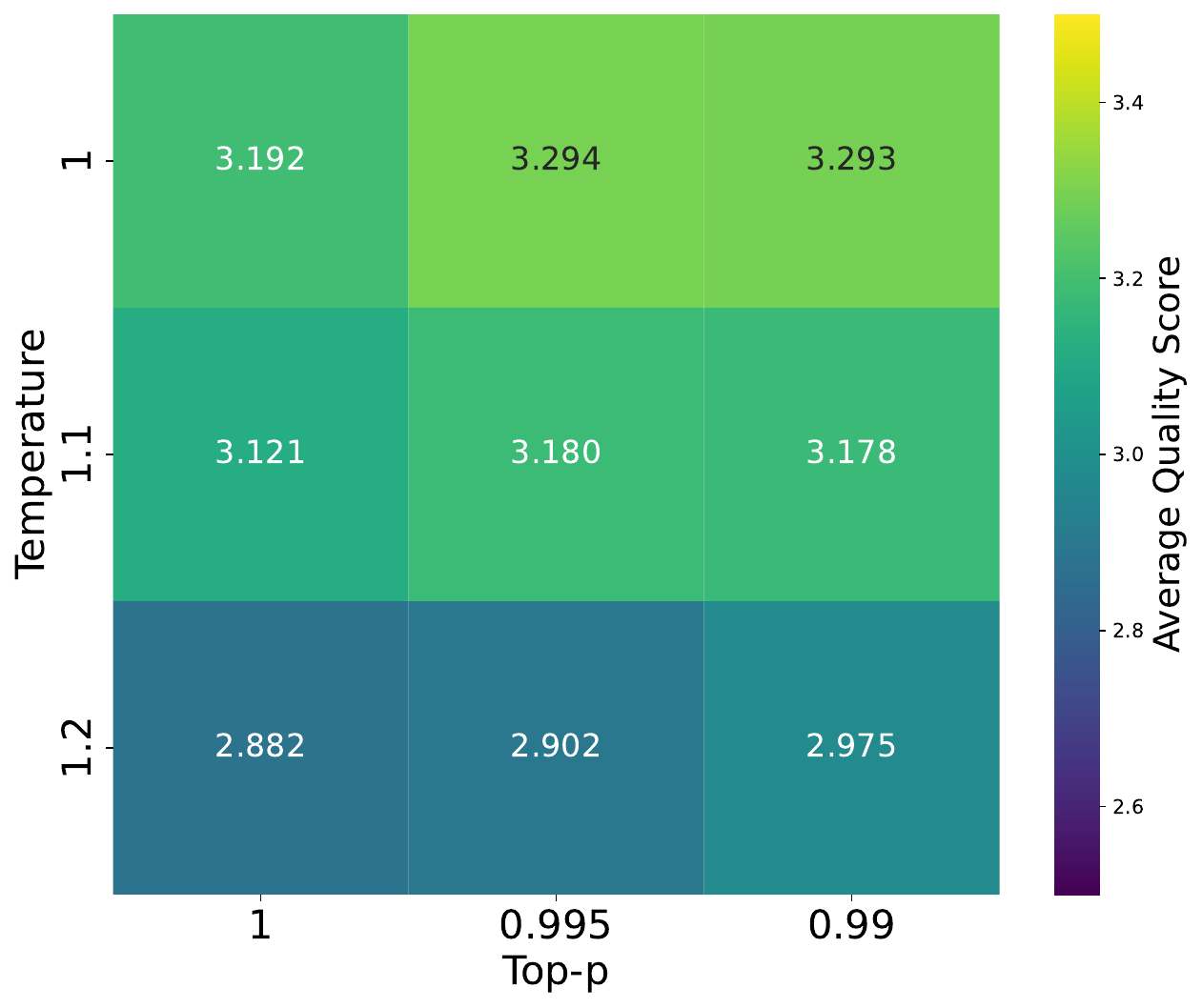}
        \caption{Average Quality Scores of \dataname{}-Air}
        \label{fig:alpha}
    \end{subfigure}
    \hfill
    \begin{subfigure}[b]{0.3\textwidth}
        \includegraphics[width=\textwidth]{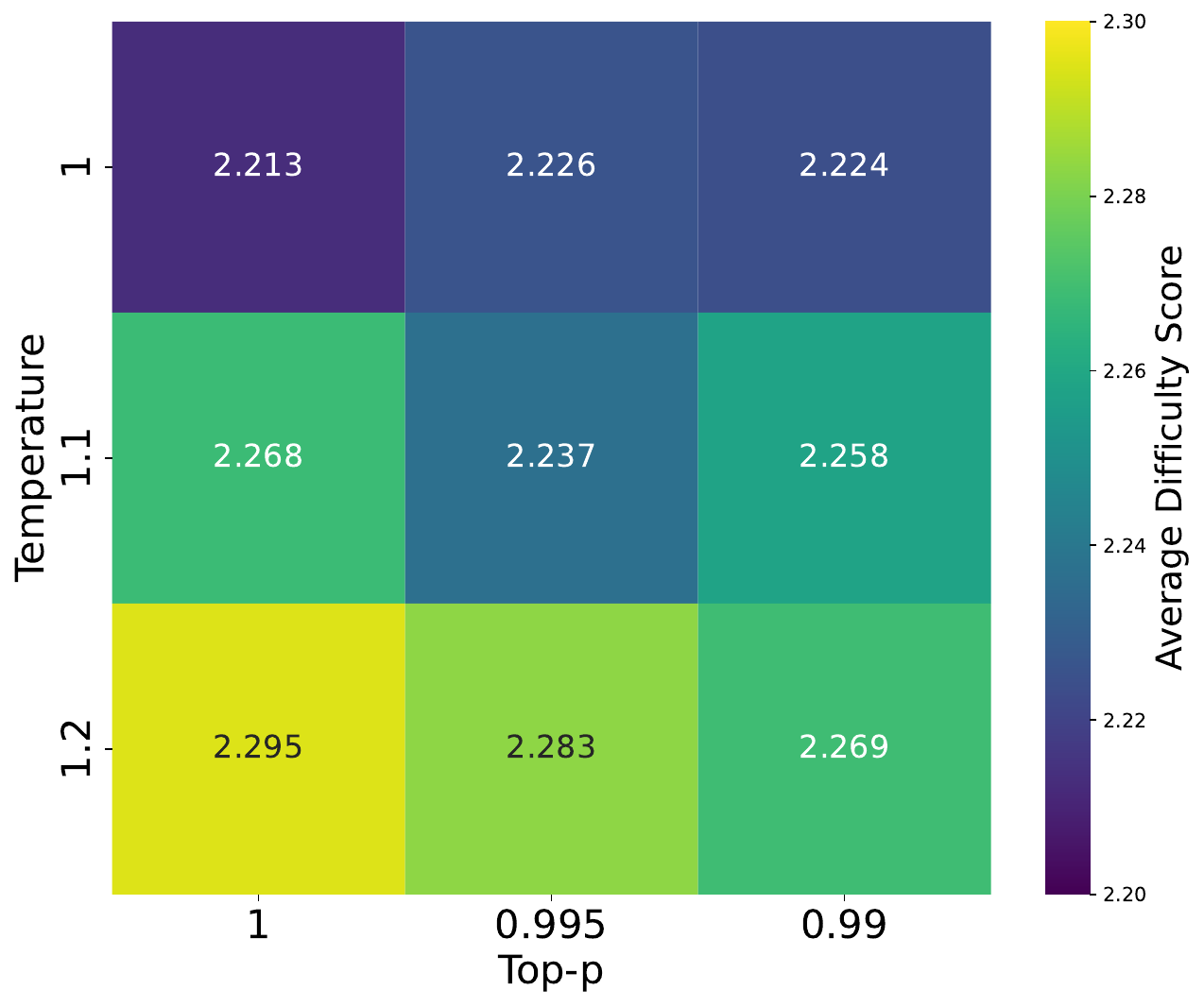}
        \caption{Average Difficulty Scores of \dataname{}-Air}
        \label{fig:m}
    \end{subfigure}
    \hfill
    \begin{subfigure}[b]{0.3\textwidth}
        \includegraphics[width=\textwidth]{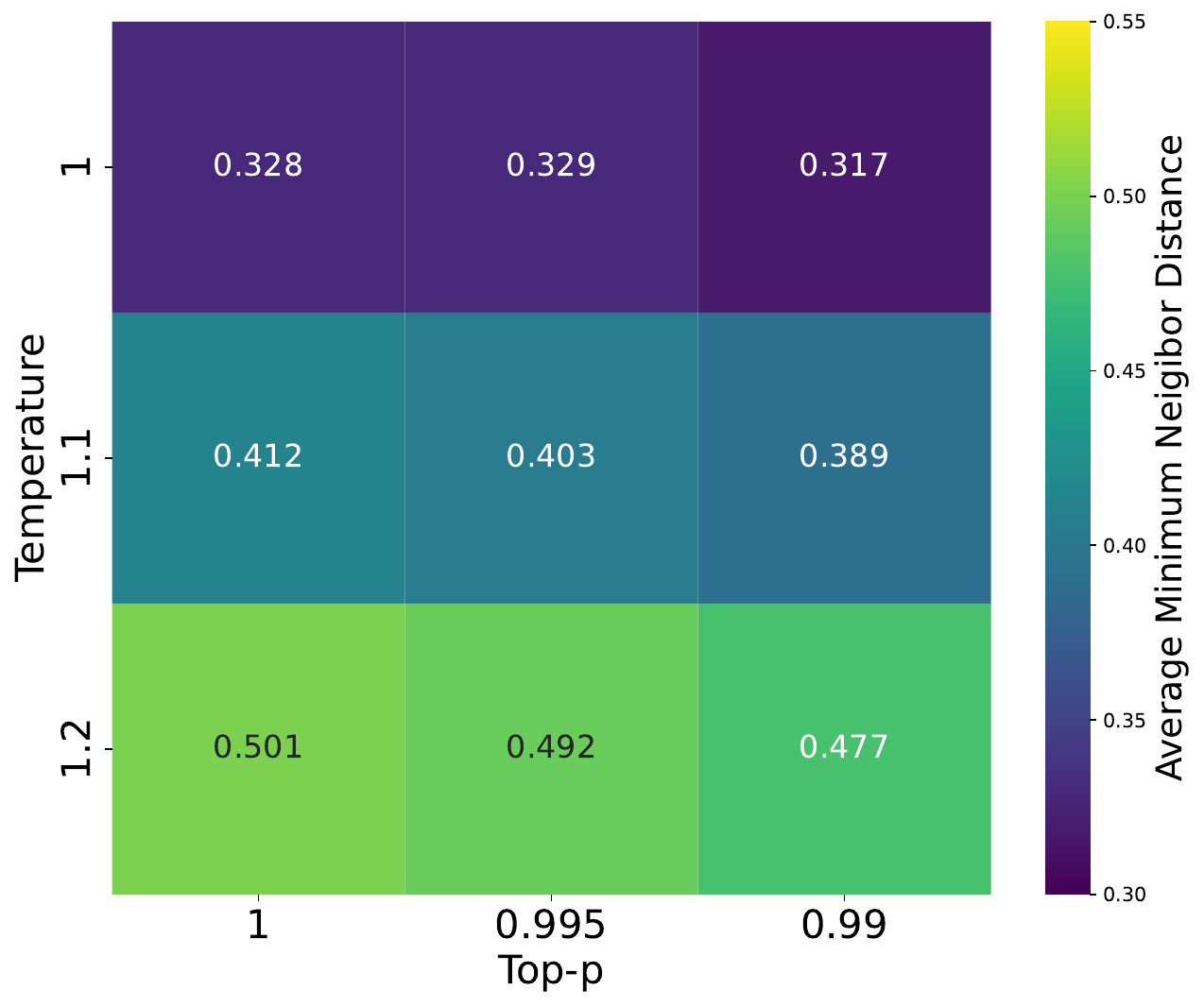}
        \caption{Average Minimum Neighbor Distances of \dataname{}-Air}
        \label{fig:c}
    \end{subfigure}
    \caption{This figure illustrates the impact of varying decoding parameters on the quality, difficulty, and diversity of generated instructions. We observe that while higher temperature and top-p values may decrease the overall quality, they tend to increase both the difficulty and diversity of the instructions.}
    \label{fig: gen config ablations}
\end{figure*}

\textbf{Ablation Analysis on Decoding Parameters.} We conduct an ablation analysis on the decoding parameters used in generating instruction with \dataname{}. Specifically, we use three different temperatures (i.e., $1$, $1.1$, and $1.2$) and top-p values (i.e., $1$, $0.995$, and $0.99$) during Step 1 of \dataname{}. We use three metrics, \textbf{Average Quality Score}, \textbf{Average Difficulty Score} and \textbf{Average Minimum Neighbor Distance} to characterize the quality, difficulty, and diversity of instructions using different decoding parameters. 
The Average Quality Score is calculated by averaging the ratings of all data within a specific temperature-top-p pair, on a scale from 1 (`very poor') to 5 (`excellent'). Similarly, the Average Difficulty Score is rated on a scale from 1 (`very easy') to 5 (`very hard').
The Average Minimum Neighbor Distance is calculated by averaging the minimum neighbor distances, as defined in Section \ref{sec:analysis}, for all data generated using the same decoding parameters.

\begin{wrapfigure}{r}{0.36\textwidth}
  \vspace{-1em}
  \centering
  \includegraphics[width=0.36\textwidth]{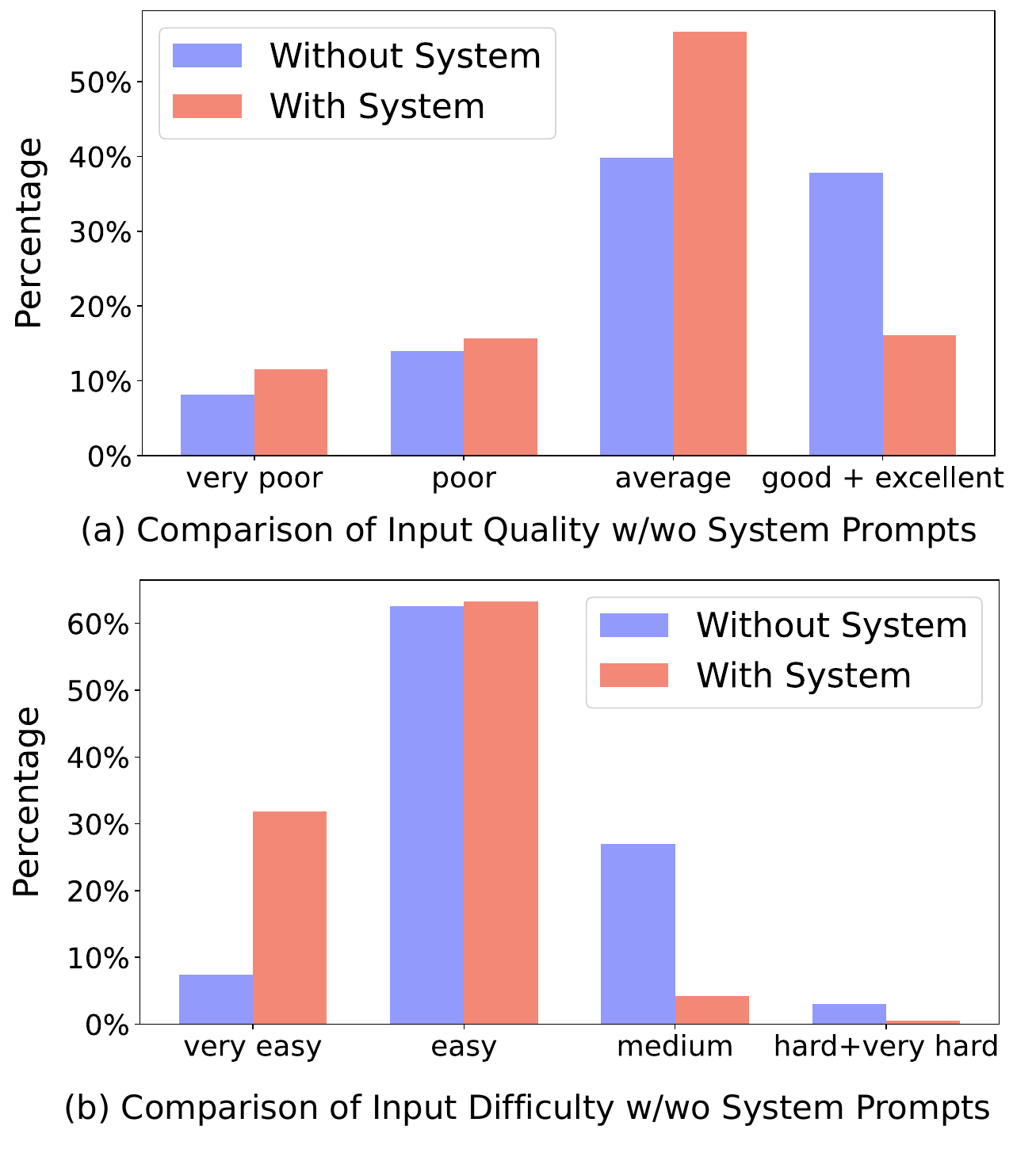}
  \caption{This figure compares the input quality and difficulty with and without system prompts.}
  \label{fig: compare system on iq}
  \vspace{-3em}
\end{wrapfigure}

The findings are summarized in Figure \ref{fig: gen config ablations}. We observe that higher temperature and top-p values may slightly decrease the overall quality of instructions, while simultaneously increasing the difficulty and remarkably enhancing the diversity of the instructions generated. The selection of these hyper-parameters should be tailored to the user's specific requirements, balancing the trade-offs between quality, difficulty, and diversity.

\textbf{Ablation Analysis on the System Prompt.} Figure \ref{fig: compare system on iq} compares the use of system prompt compared with not using it in Step 1 of \dataname{}. Since the Llama-3 model does not have an official system prompt, we use the default system prompt from Vicuna \citep{vicuna2023}: A chat between a curious user and an artificial intelligence assistant. The assistant gives helpful, detailed, and polite answers to the user's questions. We observe that using a system prompt generally results in a decrease in the overall quality of instructions, and the instructions are easier. 
Consequently, we recommend not appending system prompts in default settings.

\color{black}
\subsection{Impact of Annotating Models}
\label{appendix: impact of annotating model}
\begin{wrapfigure}{r}{0.36\textwidth}
  \centering
  \vspace{-2em}
  \includegraphics[width=0.36\textwidth]{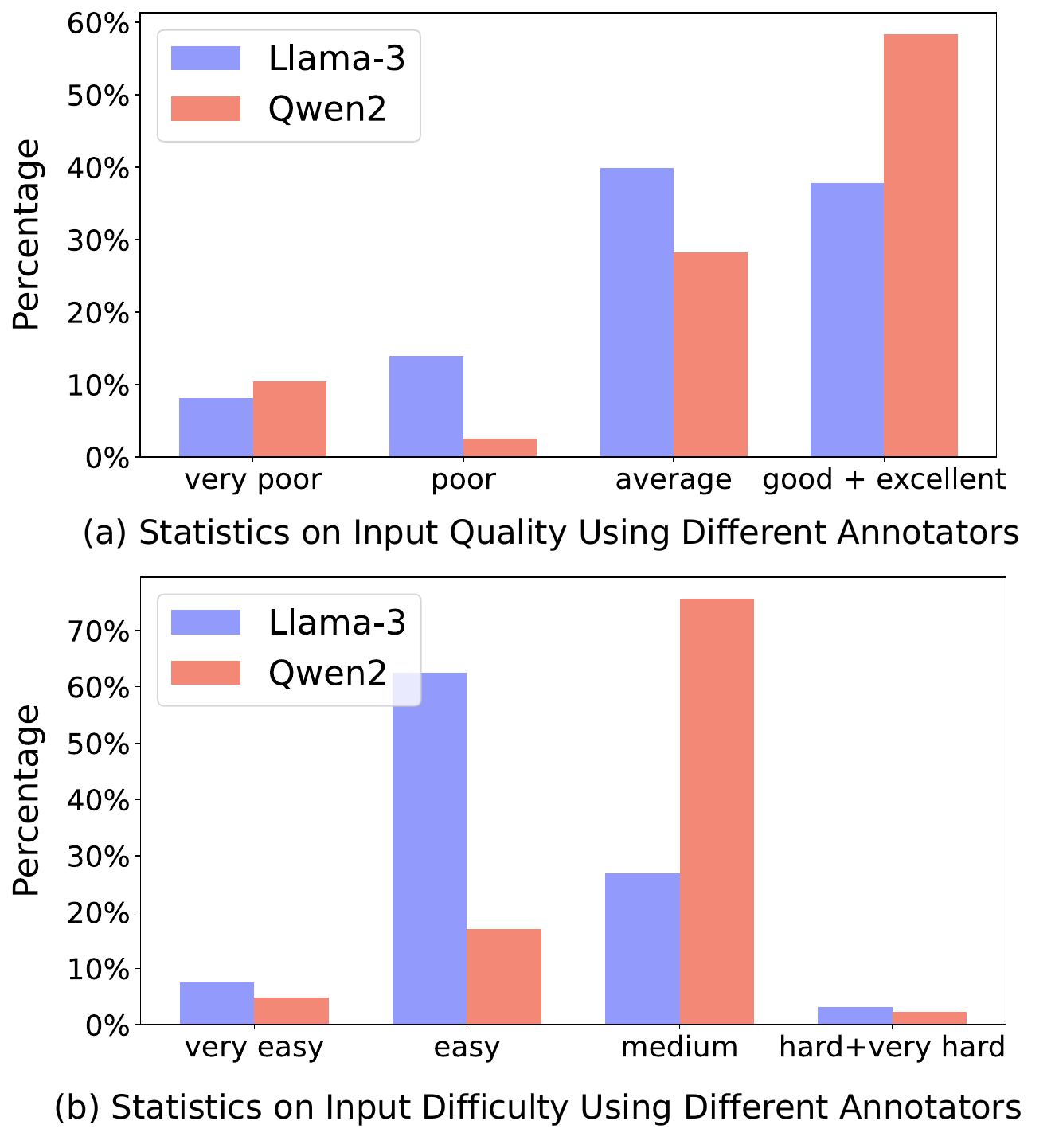}
  \caption{This figure compares the impact of different annotators on evaluating the instruction quality and difficulty.}
  \label{fig: compare annotators}
  \vspace{-5em}
\end{wrapfigure}
We note that LLMs may occasionally favor its own response \citep{deutsch-etal-2022-limitations}. In what follows, we conduct experiments to evaluate the impact of annotating models when labeling quality and difficulty of the \dataname{}-Air dataset. We used the Qwen-2-7B-Instruct model (outside the Llama-3 family) to annotate the quality and difficulty of our \dataname{}-Air dataset. The statistics are summarized in Figure \ref{fig: compare annotators}.

Our findings show that even when evaluated by Qwen-2-7B-Instruct, the \dataname{}-Air dataset maintains high quality and difficulty, which is even higher than those originally annotated by Llama-3-8B-Instruct. This suggests that our dataset's quality is robust across different annotators.

\color{black}

\section{Detailed Experimental Setups}
\label{appendix: detailed experimental setups}

\subsection{Experimental Setups for Generating \dataname{}-Air and \dataname{}-Pro}

As detailed in Appendix \ref{appendix: ablation generation config}, varying decoding parameters in Step 1 can significantly influence the quality, difficulty, and diversity of the generated instructions. To optimize the trade-offs among these attributes, we employ diverse decoding parameters for the generation of \dataname{}-Air and \dataname{}-Pro. Table \ref{tab: dataset_configs} presents the configurations of \dataname{}-Air and \dataname{}-Pro, showcasing how diverse decoding parameters shape each dataset. 

We employ greedy decoding to generate responses in Step 2 for \dataname{}-Air and \dataname{}-Pro. The intuition is that the word with the highest probability is more likely to originate from the model's training dataset.

\begin{table}[htbp]
\centering
\small
\caption{This table demonstrates the configurations of generating instructions of \dataname{}-Air and \dataname{}-Pro datasets with varying decoding parameters.}
\vspace{1em}
\begin{tabular}{c c c c c}
\toprule
\multirow{2}{*}{\textbf{Dataset}} & \multicolumn{3}{c}{\textbf{Decoding Parameters}} & \multirow{2}{*}{\textbf{Total \#Convs}} \\
\cmidrule(lr){2-4}
 & Temperature & Top-p & \#Convs \\
\midrule
\multirow{12}{*}{\dataname{}-Air} & 1.0 & 1.00 & 300K & \multirow{12}{*}{3M} \\
& 1.0 & 0.995 & 300K \\
& 1.0 & 0.990 & 300K \\
& 1.1 & 1.00 & 300K \\
& 1.1 & 0.995 & 300K \\
& 1.1 & 0.990 & 300K \\
& 1.2 & 1.00 & 300K \\
& 1.2 & 0.995 & 300K \\
& 1.2 & 0.990 & 300K \\
& 1.25 & 1.00 & 100K \\
& 1.25 & 0.995 & 100K \\
& 1.25 & 0.990 & 100K \\
      
\midrule
\multirow{4}{*}{\dataname{}-Pro} & 1.0 & 1.00 & 300K & \multirow{4}{*}{1M} \\
& 1.1 & 0.995 & 300K \\
& 1.2 & 0.995 & 300K \\
& 1.25 & 0.990 & 100K \\
\bottomrule
\end{tabular}
\label{tab: dataset_configs}
\end{table}

\subsection{Experimental Setups for Instruction Tuning and Preference Tuning}
\label{appendix: Experimental Setups for Instruction Tuning and Preference Tuning}

\textbf{Supervised Fine-Tuning Hyper-parameters.} Table \ref{tab: fine-tune hyperparameters} demonstrates the detailed supervised fine-tuning hyper-parameters. These experiments were conducted using Axolotl\footnote{\url{https://github.com/OpenAccess-AI-Collective/axolotl}}.

\begin{table}[htbp]
\small
\centering
\caption{This table shows the hyper-parameters for supervised fine-tuning.}
\vspace{1em}
\begin{tabular}{ll}
\toprule
\textbf{Hyper-parameter} & \textbf{Value} \\ \midrule
Learning Rate & $2 \times 10^{-5}$ \\
Number of Epochs & $2$ \\
Number of Devices & $4$ \\
Per-device Batch Size & $1$ \\
Gradient Accumulation Steps & $8$ \\
Effective Batch Size & $32$ \\
Optimizer & \texttt{Adamw} with $\beta s=(0.9,0.999)$ and $\epsilon=10^{-8}$\\
Learning Rate Scheduler & \texttt{cosine} \\
Warmup Steps & $100$ \\
Max Sequence Length  & $8192$ \\ \bottomrule
\end{tabular}
\label{tab: fine-tune hyperparameters}
\end{table}

\begin{table}[htbp]
\small
\centering
\caption{This table shows the hyper-parameters for direct preference optimization.}
\vspace{1em}
\begin{tabular}{ll}
\toprule
\textbf{Hyper-parameter} & \textbf{Value} \\ \midrule
Learning Rate & $5 \times 10^{-7}$ \\
Number of Epochs & $1$ \\
Number of Devices & $4$ \\
Per-device Batch Size & $2$ \\
Gradient Accumulation Steps & $16$ \\
Effective Batch Size & $128$ \\
Optimizer & \texttt{Adamw} with $\beta s=(0.9,0.999)$ and $\epsilon=10^{-8}$\\
Learning Rate Scheduler & \texttt{cosine} \\
Warmup Ratio & $10\%$ \\ \bottomrule
\end{tabular}
\label{tab: po hyperparameters}
\end{table}

\textbf{Preference Tuning Hyper-parameters.} Table \ref{tab: po hyperparameters} demonstrates the detailed DPO hyper-parameters for aligning Llama-3-8B using \dataname{}-Air-DPO and \dataname{}-Pro-DPO. These experiments were conducted using Alignment Handbook\footnote{\url{https://github.com/huggingface/alignment-handbook}}.

\textbf{Decoding parameters for evaluation benchmarks.} For Arena-Hard \citep{arenahard2024} and WildBench \citep{wildbench2024}, we follow its default setting and use greedy decoding for all settings. For AlpacaEval 2 \citep{alpaca_eval} which allows the model provider to specify decoding parameters, we also employ greedy decoding in all experiments with a slightly increased repetition penalty ($RP=1.2$) to mitigate the potential repetitive outputs during the generation.

\section{Additional Experimental Results}
\label{appendix: more exp results}

\subsection{Performance of \dataname{}-MT}
\label{appendix: multi-round magpie}

Table \ref{tab: magpie mt} compares the performance of \dataname{}-Air-MT and \dataname{}-Pro-MT with their respective single-turn counterparts. We observe that the multi-turn datasets have enhanced performance, particularly in the Arena-Hard benchmark.

\begin{table}[htbp]
    \centering
    \caption{This table compares the performance of the multi-turn versions, \dataname{}-Air-MT and \dataname{}-Pro-MT, with their single-turn counterparts. All models are instruction-tuned on the Llama-8B base models.}
    \vspace{1em}
    \resizebox{0.85\textwidth}{!}{
    \begin{tabular}{c l c c c c c
 c c c c c c c c}\toprule
    \multicolumn{2}{c}{\multirow{3}[3]{*}{\textbf{Dataset}}} & \multicolumn{6}{c}{\textbf{AlpacaEval 2}} & \textbf{Arena-Hard} \\
    & &  \multicolumn{3}{c}{GPT-4-Turbo (1106)} & \multicolumn{3}{c}{Llama-3-8B-Instruct}\\ \cmidrule(lr){3-5} \cmidrule(lr){6-8}
    & &  LC (\%) & WR (\%) & SD & LC (\%) & WR (\%) & SD & WR (\%) \\ \midrule
   \multirow{2}{*}{\dataname{}-Air} & Single-Turn & 22.66 & 23.99 & 1.24 & 49.27 & 50.80 & 1.44 & 14.9 \\
   & MT & \textbf{22.98} & \textbf{24.02} & 1.27 & \textbf{49.63} & \textbf{51.42} & 1.40 & \textbf{15.5} \\
   \midrule
   \multirow{2}{*}{\dataname{}-Pro} & Single-Turn & \textbf{25.15} & \textbf{26.50} & 1.30 & 50.52 & 52.98 & 1.43 & 18.9 \\
   & MT & 24.21 & 25.19 & 1.28 & \textbf{52.92} & \textbf{54.80} & 1.41 & \textbf{20.4} \\
    \bottomrule
    \end{tabular}
    }
    \label{tab: magpie mt}
\end{table}

\color{black}

\subsection{Compare \dataname{} and Self-Instruct using Llama-3-8B-Instruct}
\label{appendix: compare with self-instruct}

To compare the performance of \dataname{} and other synthetic dataset generation methods using the same model, we follow the official Self-Instruct \citep{wang-etal-2023-self-instruct} setup and generate a 100K supervised fine-tuning dataset using Llama-3-8B-Instruct. For a fair comparison, we select the first 100K data samples from the \dataname{}-Air dataset generated by Llama-3-8B-Instruct. The performance of models fine-tuned with these two datasets is shown in the table \ref{tab: compare with self-instruct}.

\begin{table}[bp]
    \centering
    \caption{This table compares the performance of models fine-tuned using 100K instruction-following datasets generated by Self-Instruct and \dataname{}. All models are supervised-fine-tuned on the Llama-8B base models. We observe that \dataname{} significantly outperforms Self-Instruct across all benchmarks.}
    \vspace{1em}
    \resizebox{0.85\textwidth}{!}{
    \begin{tabular}{c c c c c
 c c c c c c c c}\toprule
    \multicolumn{1}{c}{\multirow{3}[3]{*}{\textbf{Dataset}}} & \multicolumn{1}{c}{\multirow{3}[3]{*}{\#Convs}} & \multicolumn{6}{c}{\textbf{AlpacaEval 2}} & \textbf{Arena-Hard} \\
    & & \multicolumn{3}{c}{GPT-4-Turbo (1106)} & \multicolumn{3}{c}{Llama-3-8B-Instruct}\\ \cmidrule(lr){3-5} \cmidrule(lr){6-8}
    & & LC (\%) & WR (\%) & SD & LC (\%) & WR (\%) & SD & WR (\%) \\ \midrule
   \dataname{}-Air-100K & 100K & 20.17 & 21.33 & 1.21 & 46.82 & 48.76 & 1.44 & 15.7 \\
   Self-Instruct (Llama-3) & 100K & 7.21 & 5.18 & 0.7 & 17.86 & 12.73 & 1.05 & 4.0 \\
    \bottomrule
    \end{tabular}
    }
    \label{tab: compare with self-instruct}
\end{table}
We observe a significant performance gap between models fine-tuned with datasets generated by Self-Instruct and our \dataname{}. Our analysis revealed that the instruction format in Self-Instruct-generated datasets is predominantly constrained by the patterns defined in the seed instructions, resulting in a lack of diversity. This comparison indicates the novelty of our \dataname{} in generating diverse high-quality instructions without any seed questions.

\subsection{Ablation Analysis on Data Quantity and Quality}
\label{appendix: data quantity and quality}
\color{black}
In what follows, we compare within the family of datasets generated by \dataname{} in Table \ref{tab:amount}.
These datasets differ in size, deployment of filtering, and models used to generate data.
We observe that as the dataset's size increases, the fine-tuned model's performance improves, indicating that data quantity plays a critical role in enhancing instruction-following capabilities.
Furthermore, the model fine-tuned with \dataname{}-Pro-300K-Filtered outperforms those fine-tuned with the same or even higher amounts of raw data.
This demonstrates the effectiveness of our filtering technique, and underscores the importance of data quality.
Finally, we observe that the models fine-tuned with \dataname{}-Pro consistently outperform those fine-tuned with \dataname{}-Air. The reason is that  \dataname{}-Pro is generated by the more capable model, i.e., Llama-3-70B-Instruct.

\begin{table}[htbp]
    \centering
    \caption{\textcolor{black}{This table compares \dataname{} datasets within its family that differ in size, deployment of filtering, and models used to generate data. All models are supervised-fine-tuned on the Llama-8B base models.}}
    \vspace{1em}
    \resizebox{\textwidth}{!}{
    \begin{tabular}{c l   c   c c c c c c
 c c c c}\toprule
    \multicolumn{2}{c}{\multirow{3}[3]{*}{\makecell{\textbf{Dataset}}}}   & \multirow{3}[3]{*}{\makecell{\textbf{\#Convs}}} & \multicolumn{6}{c}{\textbf{AlpacaEval 2}} & \textbf{Arena-Hard} \\ 
    & & &  \multicolumn{3}{c}{GPT-4-Turbo (1106)} & \multicolumn{3}{c}{Llama-3-8B-Instruct} & \\ \cmidrule(lr){4-6} \cmidrule(lr){7-9}
   &  & & LC (\%) & WR (\%) & SD & LC (\%) & WR (\%) & SD & WR(\%) \\ \midrule

     \multirow{3}{*}{\dataname{}-Air} & \multicolumn{1}{l}{300K-Raw}  & 300K & 21.99 & 21.65 & 1.21 & 48.63 & 48.06 & 1.42 & 15.8\\
     & \multicolumn{1}{l}{3M-Raw}  & 3M & 22.96 & 21.09 & 1.20 & 50.57 & 48.40 & 1.42 & 16.1\\
     & \multicolumn{1}{l}{300K-Filtered} & 300K & 22.66 & {23.99} & 1.24 & 49.27 & {50.8} & 1.44 & 14.9 \\  \midrule
     \multirow{5}{*}{\dataname{}-Pro} & \multicolumn{1}{l}{300K-Raw}  & 300K & 21.65 & 22.19 & 1.2 & 49.65 & {50.84} & 1.42 & 15.9 \\
     & \multicolumn{1}{l}{1M-Raw} & 1M & 24.16 & 23.93 & 1.25 & 49.97 & {50.34} & 1.43 & 16.7 \\
     & \multicolumn{1}{l}{100K-Filtered}  & 100K & 20.47 & {24.52} & 1.25 & 47.92 & {52.75} & 1.43 & 17.2
     \\
     & \multicolumn{1}{l}{200K-Filtered}  & 200K & 22.11 & {26.02} & 1.26 & {51.17} & \textbf{56.76} & 1.41 & 15.9
     \\
     & \multicolumn{1}{l}{300K-Filtered} & 300K & \textbf{25.08} & \textbf{29.47} & 1.35 & \textbf{52.12} & {53.43} & 1.44 & \textbf{18.9} \\ \midrule
     \multirow{1}{*}{\dataname{}-Air + \dataname{}-Pro} & \multicolumn{1}{l}{4M-Raw} & 4M & 24.45 & 24.08 & 1.26 & 51.96 & 52.08 & 1.42 & 15.5 \\

     \bottomrule
    \end{tabular}}
    \vspace{-1em} 
    \label{tab:amount}
\end{table}

\color{black}

\subsection{Ablation Analysis on Filter Designs}
\label{appendix: ablation on filter design}

We conduct an ablation analysis on various filter designs within \dataname{}-Pro to assess their impact on the performance of supervised fine-tuned models. The results are presented in Table \ref{tab: ablation on different filter design}. We observe that different filtering strategies yield optimal performance on different benchmarks, and no single filter consistently achieves the best performance across all benchmarks. Therefore, determining how to select instructional data to enhance the performance in supervised fine-tuning is an interesting topic for future research.

\begin{table}[htbp]
    \centering
    \caption{This table compares the performance of different filter designs within \dataname{}-Pro. All models are supervised-fine-tuned on the Llama-8B base models.}
    \vspace{1em}
    \resizebox{0.8\textwidth}{!}{
    \begin{tabular}{c l c c c c c c c c c c c c c}\toprule
    \multicolumn{2}{c}{\multirow{3}[3]{*}{\textbf{Dataset and Filter}}} & \multicolumn{6}{c}{\textbf{AlpacaEval 2}} & \textbf{Arena-Hard} \\
    & &  \multicolumn{3}{c}{GPT-4-Turbo (1106)} & \multicolumn{3}{c}{Llama-3-8B-Instruct}\\ \cmidrule(lr){3-5} \cmidrule(lr){6-8}
    & &  LC (\%) & WR (\%) & SD & LC (\%) & WR (\%) & SD & WR (\%) \\ \midrule
   \multirow{6}{*}{\dataname{}-Pro} & Filter
   & 25.08 & \textbf{29.47} & 1.35 & 52.12 & 53.43 & 1.44 & \textbf{18.9} \\
   & Filter 2 & \textbf{25.15} & 26.50 & 1.30 & 50.52 & 52.98 & 1.43 & \textbf{18.9} \\
   & Filter 3 & 23.90 & 25.21 & 1.25 & 51.45 & 53.64 & 1.41 & 18.3 \\
   & Filter 4 & 24.20 & 25.33 & 1.27 & \textbf{52.43} & 54.34 & 1.43 & 17.9 \\
   & Filter 5 & 24.85 & 25.12 & 1.26 & 52.12 & 53.43 & 1.44 & 18.4 \\
   & Filter 6 & 23.20 & 28.43 & 1.26 & 51.34 & \textbf{57.29} & 1.41 & 17.9 \\
    \bottomrule
    \end{tabular}
    }
    \label{tab: ablation on different filter design}
\end{table}

\color{black}
\subsection{Ablation Analysis on Response Generator}
\label{appendix: ablation on output generator}

To investigate the impact of the response generator on the supervised fine-tuning performance using \dataname{}, we conduct an ablation study by replacing the response generator with Qwen-2-7B-Instruct \citep{yang2024qwen2} within \dataname{}-Air-300K-Filtered. We note that the performance of Qwen-2-7B-Instruct is comparable to, or slightly weaker than, Llama-3-8B-Instruct. The results are summarized in Table \ref{tab: magpie different generator}.

We observe that although there is a slight performance degradation, the model fine-tuned using Qwen-2-7B-Instruct as the response generator still outperforms all baselines, including those using GPT-4 as the response generator. These findings indicate two key points: (1) The success of \dataname{} depends little on the specific response generator used, and (2) the instructions generated by \dataname{} are of high quality and diversity.

\begin{table}[htbp]
    \centering
    \caption{This table compares the impact of different response generators on the model performance. All models are supervised-fine-tuned on the Llama-8B base models.}
    \resizebox{0.85\textwidth}{!}{
    \begin{tabular}{c c c c c
 c c c c c c c c}\toprule
    \multicolumn{1}{c}{\multirow{3}[3]{*}{\textbf{Response Generator}}} & \multicolumn{6}{c}{\textbf{AlpacaEval 2}} & \textbf{Arena-Hard} \\
    & \multicolumn{3}{c}{GPT-4-Turbo (1106)} & \multicolumn{3}{c}{Llama-3-8B-Instruct}\\ \cmidrule(lr){2-4} \cmidrule(lr){5-7}
    & LC (\%) & WR (\%) & SD & LC (\%) & WR (\%) & SD & WR (\%) \\ \midrule
   Llama-3-8B-Instruct & 22.66 & 23.99 & 1.24 & 49.27 & 50.80 & 1.44 & 14.9 \\
   Qwen2-7B-Instruct & 15.01 & 15.60 & 1.05 & 41.09 & 42.07 & 1.47 & 13.7 \\
    \bottomrule
    \end{tabular}
    }
    \label{tab: magpie different generator}
\end{table}

\subsection{Trustworthiness of \dataname{}-Aligned Models}
\label{appendix: compare with official}


In what follows, we conduct more experiments to compare \dataname{} model and Llama-3-8B-Instruct on the TrustLLM benchmark \citep{huang2024trustllm}. The results for safety, fairness, ethics, privacy, and robustness are summarized in Table \ref{tab:vs-llama3official}.

We observe that our supervised-fine-tuned model slightly underperforms Llama-3-8B-Instruct in terms of safety and fairness. However, it outperforms the official instruct model on ethics, privacy, and robustness. Considering that our fine-tuned model uses much fewer data samples (300K compared to over 10M), these results again highlight the high quality of data generated by \dataname{}.

\begin{table}[htbp]
    \centering
    \caption{This table compares the performance of model supervised-fine-tuned using \dataname{}-Pro-300K-Filtered and the official Llama-3-8B-Instruct on the TrustLLM benchmark \citep{huang2024trustllm}.}
    \resizebox{\textwidth}{!}{
    \begin{tabular}{ c c c c c c }
    \toprule
        \multirow{1}{*}{\makecell{\textbf{TrustLLM}}} & Evaluation/Dataset & Llama-3-8B-Instruct & \dataname{}-Pro-300K-Filtered \\ \midrule
        \multirow{3}{*}{Safety} & Jailbreak (\small{RtA$\uparrow$}) & \bf 0.93 & 0.80 \\
        & Misuse (\small{RtA$\uparrow$}) & \bf 0.85 & 0.80 \\
        & Exaggerated Safety (\small{RtA$\downarrow$}) & \bf 0.54 & 0.52 \\ \midrule
        \multirow{4}{*}{Fairness} & Stereotype Recognition (\small{Acc$\uparrow$}) & \bf 0.49 & 0.40 \\
        & Stereotype Query Test (\small{RtA$\uparrow$}) & \bf 1.00 & 0.99 \\
        & Disparagement Sex (\small{p-value$\uparrow$}) & \bf 0.99 & \bf 0.99\\
        & Disparagement Race (\small{p-value$\uparrow$}) & \bf 0.55 & 0.47 \\ \midrule
        \multirow{4}{*}{Ethics}  & Social Chemistry 101 (\small{Acc$\uparrow$}) & \bf 0.94 & 0.63 \\
        & ETHICS (\small{Acc$\uparrow$}) & 0.65 & \bf 0.69 \\
        & MoralChoice (\small{Acc$\uparrow$}) & \bf 0.97 & 0.95\\
        & MoralChoice (\small{RtA$\uparrow$}) & 0.97 & \bf 0.98 \\ \midrule
        \multirow{3}{*}{Privacy} & Privacy Awareness-Normal (\small{RtA$\uparrow$}) & 0.33 & \bf 0.71 \\
        & Privacy Awareness-Augmented (\small{RtA$\uparrow$}) & \bf 1.00& 0.98 \\
        & Privacy Leakage (\small{RtA$\uparrow$}) & 0.66 & \bf 0.87 \\ \midrule
        \multirow{3}{*}{Robustness} & AdvGlue (\small{RobustScore$\uparrow$})  & 0.42 & \bf 0.58 \\
        & OOD Detection (\small{RtA$\uparrow$}) &  \bf 0.37 & 0.26\\
        & OOD Generalization (\small{F1-Score$\uparrow$}) & 0.83 & \bf 0.84\\ \bottomrule
    \end{tabular}
    }
    \label{tab:vs-llama3official}
\end{table}

\section{Prompt Templates}
\label{appendix: prompt template}

\subsection{Prompt Templates for \dataname{} Extension}

This section presents the prompt template used to generate \dataname{}-MT and control instruction tasks, as detailed in Figure \ref{fig: generating mt prompt} and Figure \ref{fig: control generation topic}, respectively.

\begin{figure}[htbp]
    \centering
\begin{tcolorbox}[title=Prompt for generating \dataname{}-MT, promptstyle]
\lstset{
    basicstyle=\normalfont\sffamily\footnotesize,
    breaklines=true,
    frame=none,
    columns=fullflexible,
}
\begin{lstlisting}
<|begin_of_text|><|start_header_id|>system<|end_header_id|>

You are a helpful Al assistant. The user will engage in a multi-round conversation with you, asking initial questions and following up with additional related questions. Your goal is to provide thorough, relevant and insightful responses to help the user with their queries.<|eot_id|><|start_header_id|>user<|end_header_id|>

{instruction}<|eot_id|><|start_header_id|>assistant<|end_header_id|>

{response}<|eot_id|><|start_header_id|>user<|end_header_id|>

\end{lstlisting}
\end{tcolorbox}
    \caption{Prompt for generating \dataname{}-MT. We take Llama-3-8B-Instruct as an example. The placeholder \texttt{\{instruction\}} and \texttt{\{response\}} are from the first turn.}
    \label{fig: generating mt prompt}
\end{figure}

\begin{figure}
    \centering

\begin{tcolorbox}[title=System Prompt Template, promptstyle]
\lstset{
    basicstyle=\normalfont\sffamily\footnotesize,
    breaklines=true,
    frame=none,
    columns=fullflexible,
}
\begin{lstlisting}
<|begin_of_text|><|start_header_id|>system<|end_header_id|>

{System Prompt}<|eot_id|><|start_header_id|>user<|end_header_id|>

\end{lstlisting}
\end{tcolorbox}

\begin{tcolorbox}[title=System prompt for controlling math instruction tasks, promptstyle]

You are an AI assistant designed to provide helpful, step-by-step guidance on solving math problems. The user will ask you a wide range of complex mathematical questions. Your purpose is to assist users in understanding mathematical concepts, working through equations, and arriving at the correct solutions.

\end{tcolorbox}

\begin{tcolorbox}[title=System prompt for controlling code instruction tasks, promptstyle]

You are an AI assistant designed to provide helpful, step-by-step guidance on coding problems. The user will ask you a wide range of coding questions. Your purpose is to assist users in understanding coding concepts, working through code, and arriving at the correct solutions.

\end{tcolorbox}

\begin{tcolorbox}[title=System prompt for controlling translation tasks, promptstyle]

You are an AI assistant designed to provide accurate and contextually appropriate translations. Users will ask you to translate text between various languages. Your purpose is to assist users in understanding and conveying meaning across languages, maintaining the original context and nuances.

\end{tcolorbox}

\begin{CJK*}{UTF8}{gbsn}
\begin{tcolorbox}[title=System prompt for controlling multilingual instruction generation (Japanese + Math), promptstyle]\

あなたはAIアシスタントで、数学の問題を解くために役立つ、ステップバイステップのガイダンスを提供するように設計されています。

\end{tcolorbox}
\end{CJK*}

\caption{Prompts for controlling instruction generation tasks. These examples illustrate how to guide Llama-3-8B-Instruct in generating instructions for specific domains: mathematics, coding, translation, and multilingual tasks. To adapt this approach for different instruction tasks, replace the \texttt{{System Prompt}} placeholder in the System Prompt Template with the appropriate domain-specific prompt.}
\label{fig: control generation topic}
\end{figure}

\subsection{Prompt Templates for Evaluation}

Here, we present the prompt template employed to generate task categories, quality, and difficulty, as detailed in Figure \ref{fig: task category prompt}, Figure \ref{fig: quality prompt}, and Figure \ref{fig: difficulty prompt}, respectively. The placeholder \texttt{{input}} represents the instructions to be evaluated.

\begin{figure}
\centering
\begin{tcolorbox}[title=Prompt for generating task categories, promptstyle]
\lstset{
    basicstyle=\normalfont\sffamily\footnotesize,
    breaklines=true,
    frame=none,
    columns=fullflexible,
}
\begin{lstlisting}
# Instruction
Please label the task tags for the user query.

## User Query
```{input}```

## Tagging the user input
Please label the task tags for the user query. You will need to analyze the user query and select the most relevant task tag from the list below.

all_task_tags = [
    "Information seeking",  # Users ask for specific information or facts about various topics.
    "Reasoning",  # Queries require logical thinking, problem-solving, or processing of complex ideas.
    "Planning",  # Users need assistance in creating plans or strategies for activities and projects.
    "Editing",  # Involves editing, rephrasing, proofreading, or other tasks related to the composition of general written content.
    "Coding & Debugging",  # Users seek help with writing, reviewing, or fixing code in programming.
    "Math",  # Queries related to mathematical concepts, problems, and calculations.
    "Role playing",  # Users engage in scenarios requiring ChatGPT to adopt a character or persona.
    "Data analysis",  # Requests involve interpreting data, statistics, or performing analytical tasks.
    "Creative writing",  # Users seek assistance with crafting stories, poems, or other creative texts. 
    "Advice seeking",  # Users ask for recommendations or guidance on various personal or professional issues.
    "Brainstorming",  # Involves generating ideas, creative thinking, or exploring possibilities. 
    "Others"  # Any queries that do not fit into the above categories or are of a miscellaneous nature.
]

## Output Format:
Note that you can only select a single primary tag. Other applicable tags can be added to the list of other tags. 
Now, please output your tags below in a json format by filling in the placeholders in <...>:
```
{{ 
    "primary_tag": "<primary tag>",
    "other_tags": ["<tag 1>", "<tag 2>", ... ]
}}
```
\end{lstlisting}
\end{tcolorbox}
    \caption{Prompt for generating task categories}
    \label{fig: task category prompt}
\end{figure}

\begin{figure}
    \centering
\begin{tcolorbox}[title=Prompt for generating quality of instructions, promptstyle]
\lstset{
    basicstyle=\normalfont\sffamily\footnotesize,
    breaklines=true,
    frame=none,
    columns=fullflexible,
}
\begin{lstlisting}
# Instruction
You need to rate the quality of the user query based on its clarity, specificity, and coherence.
The rating scale is as follows:

- very poor: The query is unclear, vague, or incoherent. It lacks essential information and context.
- poor: The query is somewhat unclear or lacks important details. It requires significant clarification.
- average: The query is moderately clear and specific. It may require some additional information for a complete understanding.
- good: The query is clear, specific, and mostly well-formed. It provides sufficient context for understanding the user's intent.
- excellent: The query is very clear, specific, and well-articulated. It contains all the necessary information and context for providing a comprehensive response.

## User Query
```{input}```

## Output Format
Given the user query, you first need to give an assessment, highlighting the strengths and/or weaknesses of the user query. Then, you need to output a rating from very poor to excellent by filling in the placeholders in [...]:
```
{{   
    "explanation": "[...]",
    "input_quality": "[very poor/poor/average/good/excellent]"
}}
```
'''
\end{lstlisting}
\end{tcolorbox}
    \caption{Prompt for generating quality of instructions}
    \label{fig: quality prompt}
\end{figure}

\begin{figure}
    \centering
\begin{tcolorbox}[title=Prompt for generating difficulty of instructions, promptstyle]
\lstset{
    basicstyle=\normalfont\sffamily\footnotesize,
    breaklines=true,
    frame=none,
    columns=fullflexible,
}
\begin{lstlisting}
# Instruction 
You first need to identify the given user intent and then label the difficulty level of the user query based on the content of the user query.

## User Query
```{input}```

## Output Format
Given the user query, in your output, you first need to identify the user intent and the knowledge needed to solve the task in the user query. Then, rate the difficulty level of the user query as `very easy`, `easy`, `medium`, `hard`, or `very hard`.

Now, please output the user intent and difficulty level below in a json format by filling in the placeholders in [...]:
```
{{   
    "intent": "The user wants to [....]",
    "knowledge": "To solve this problem, the models need to know [....]",
    "difficulty": "[very easy/easy/medium/hard/very hard]"
}}
```
\end{lstlisting}
\end{tcolorbox}
    \caption{Prompt for generating difficulty of instructions}
    \label{fig: difficulty prompt}
\end{figure}

\clearpage
\section{\dataname{} Examples}
\label{appendix: magpie examples}

In this section, we demonstrate several representative instructions generated by \dataname{} with different task categories. The following instructions are generated by the Llama-3-70B-Instruct model.

\begin{tcolorbox}[title=Information Seeking, promptstyle]
A few days ago, I was at a restaurant and I got a cup of coffee. However, when I went to take a sip, I realized it was a little too hot. So, I decided to let it cool down for a few minutes. As I waited, I noticed that coffee is actually two different colors. The part that was closest to the surface of the coffee is a lighter color, and the part that is deeper is a darker color. Have you ever observed this phenomenon before?
\end{tcolorbox}

\begin{tcolorbox}[title=Reasoning, promptstyle]
You have 3 switches but they are not labeled. Each switch corresponds to one of three light bulbs in a room. Each light bulb is either on or off. You can turn the switches on and off as many times as you want, but you can only enter the room one time to observe the bulbs. How can you figure out which switch corresponds to which light bulb?
\end{tcolorbox}

\begin{tcolorbox}[title=Planning, promptstyle]
You are the Founder of a Financial Planning Company. As a professional financial advisor, you are scheduled to meet a new client tomorrow. Specifically, what are you planning to do to prepare for this meeting?
\end{tcolorbox}

\begin{tcolorbox}[title=Editing, promptstyle]
What is the best way to re-write the sentence: "We call this the `core' product or the `core' offering" using proper quotation marks and avoiding the word "this"?
\end{tcolorbox}

\begin{tcolorbox}[title=Coding \& Debugging, promptstyle]
Write a Python program that calculates the total cost of a customer's order. The program should ask for the customer's name, the number of items they want to purchase, and the price of each item. It should then calculate the total cost by multiplying the number of items by the price of each item and adding 8\% sales tax. The program should display the customer's name, the number of items, the price of each item, and the total cost, including sales tax.
\end{tcolorbox}

\begin{tcolorbox}[title=Math, promptstyle]
In the following problem, please use integers to solve it. A water tank has 1000 L of water. On the first day, 1/5 of the water was drained. On the second day, 3/10 of the remaining water was drained. On the third day, 2/5 of the remaining water was drained. On the fourth day, 3/4 of the remaining water was drained. How many liters of water are left after the fourth day?
\end{tcolorbox}

\begin{tcolorbox}[title=Role Playing, promptstyle]
In this game, you will be the host, and I will be the contestant. You will ask me a series of questions, and I will try to answer them correctly. The questions will be multiple choice, and I will have a 25\% chance of getting the correct answer if I just randomly guess. However, I am a clever contestant, and I will try to use logic and reasoning to increase my chances of getting the correct answer.
\end{tcolorbox}

\begin{tcolorbox}[title=Data Analysis, promptstyle]
The personnel manager at a company is tasked with finding the average salary of new hires. She has collected data on the salaries of 13 new hires. She wants to know if there is a statistical difference between the average salary of new hires and the national average salary. The national average salary is \$45,000. The sample of new hires has a mean salary of \$42,800 and a standard deviation of \$4,200.
\end{tcolorbox}

\begin{tcolorbox}[title=Creative Writing, promptstyle]
Write a paragraph about a mythical creature that you created. The creature is small, no larger than a house cat. It has shimmering scales that reflect light, and can emit a soft, pulsing glow from its body. It has large, round eyes that seem to see right through you, but with a gentle kindness. It has a soft, melodious voice, and can communicate with humans through a form of telepathy.
\end{tcolorbox}

\begin{tcolorbox}[title=Advice Seeking, promptstyle]
How do you handle stress and overwhelm?
\end{tcolorbox}

\begin{tcolorbox}[title=Brainstorming, promptstyle]
Can you give me some ideas for a spontaneous, fun and memorable birthday celebration for my partner?
\end{tcolorbox}

\begin{tcolorbox}[title=Others, promptstyle]
What does "sdrawkcaB" mean?
\end{tcolorbox}

\vspace{1em}

\color{black}

\dataname{} can also generate domain-specific instructions using models that are tailored to particular fields, as mentioned in Section \ref{sec: magpie extension}. The following instructions are generated by DeepSeek-Coder-V2 \citep{zhu2024deepseek} and Qwen2-Math-7B-Instruct \citep{yang2024qwen2}, respectively.

\begin{tcolorbox}[title=DeepSeek-Coder-V2 (Code Instruction), promptstyle]
You are given a list of emails. You need to write a Python function that returns the domain, excluding the @ symbol, for each email.
\end{tcolorbox}

\begin{tcolorbox}[title=Qwen2-Math-7B-Instruct (Math Instruction), promptstyle]
A rectangle with length 12 units and width 8 units is scaled by a factor of 2 to form a new rectangle. Determine the dimensions of the new rectangle and calculate its area. Compare the area of the new rectangle to the area of the original rectangle.
\end{tcolorbox}

We note that \dataname{}'s capabilities extend beyond generating English datasets to producing diverse multilingual datasets. The following instructions are generated by the Qwen2-72B-Instruct model.

\begin{CJK*}{UTF8}{gbsn}

\begin{tcolorbox}[title=Chinese, promptstyle]
从给定的两个整数中找到较大的一个。但是你不能使用任何比较操作符（如>, <, !=等）或数学运算符（如+, -, *, /等）来实现它。你只能使用位操作符和逻辑操作符。
\end{tcolorbox}

\end{CJK*}

\begin{tcolorbox}[title=German, promptstyle]
Das Debate-System 'Oxford-Oberhaus' wird bei ersten Auseinandersetzungen verwendet. Bitte erklären sie, wie dieses System funktioniert.
\end{tcolorbox}

\begin{tcolorbox}[title=Spanish, promptstyle]
Según la encuesta anual de satisfacción al cliente que acabamos de realizar, parece que la satisfacción general de los clientes con nuestro rendimiento ha disminuido. ¿Podrías preparar una presentación detallada para la reunión del lunes que analice los resultados, identifique las áreas problemáticas y proporcione posibles soluciones basadas en los datos recogidos?
\end{tcolorbox}

\begin{tcolorbox}[title=Portuguese, promptstyle]
Ho comprato una nuova inchiostriera sulla quale è presente la scritta "Non manipolare". Cosa signidica?
\end{tcolorbox}

\begin{tcolorbox}[title=Italian, promptstyle]
Crie um exemplo de uma conversa entre dois personagens, um MC de hip hop e um pianista clássico, discutindo sobre seus estilos favoritos de música.
\end{tcolorbox}

\color{black}

\end{document}